\pdfoutput=1

\documentclass[11pt]{article}

\usepackage{acl}

\usepackage{times}
\usepackage{latexsym}

\usepackage[T1]{fontenc}

\usepackage[utf8]{inputenc}

\usepackage{microtype}

\usepackage{hyperref}       
\usepackage{url}            
\usepackage{booktabs}       
\usepackage{amsfonts}       
\usepackage{nicefrac}       
\usepackage{microtype}      
\usepackage{xcolor}         
\usepackage{amssymb}
\usepackage[super]{nth}
\usepackage{mathtools}
\usepackage{multirow}
\usepackage{amsmath}
\usepackage{amsthm}
\usepackage{subfig}
\usepackage{wrapfig}

\usepackage{todonotes}   
\usepackage{tikz}
\usepackage{pgfplots}

\usepackage{bm}

\usepackage{stmaryrd}  

\usepackage{tipa}

\newcommand{\Real}{\mathbb{R}}

\newcommand{\Lc}{\mathcal{L}}

\newtheorem{theorem}{Theorem}
\newtheorem{Lemma}{Lemma}
\newtheorem{Definition}{Definition}
\newtheorem{Proof}{Proof}

\usepackage{hyperref}
\usepackage[noabbrev,capitalize]{cleveref}

\crefname{equation}{equation}{equations}   
\crefname{footnote}{footnote}{footnotes}   
\crefname{Lemma}{Lemma}{Lemmas}
\crefname{section}{\S}{\S\S}
\Crefname{section}{\S}{\S\S}    
\crefformat{section}{#2\S#1#3}  
\Crefformat{section}{#2\S#1#3}
\crefrangeformat{section}{\S\S#3#1#4--#5#2#6}
\Crefrangeformat{section}{\S\S#3#1#4--#5#2#6}
\crefformat{section}{#2\S#1#3}  
\crefrangeformat{section}{\S\S#3#1#4--#5#2#6}

\newcommand{\poincare}{Poincar\'e }
\newcommand{\lorentz}{\mathbb{L}^n_K}
\newcommand{\linner}[2]{\langle \mathbf{#1},\mathbf{#2}\rangle_\Lc}
\newcommand{\lnorm}[1]{\lVert #1\rVert_\Lc}
\newcommand{\tangent}[1]{\mathcal{T}_{\mathbf{#1}}\mathbb{L}^n_K}

\newcommand{\logmap}[2]{\log_#1^K(#2)}
\newcommand{\expmap}[2]{\exp_\mathbf{#1}^K(\mathbf{#2})}
\newcommand{\riemanntensor}[1][x]{\mathfrak{g}_\mathbf{x}^K}

\newcommand*\samethanks[1][\value{footnote}]{\footnotemark[#1]}

\newtheoremstyle{named}{}{}{\itshape}{}{\bfseries}{.}{.5em}{\thmnote{#3's }#1}
\theoremstyle{named}

\DeclareMathOperator*{\argmin}{arg\,min}

\title{Fully Hyperbolic Neural Networks}

\author{Weize Chen$^{1,3}$\thanks{\ \ Equal contribution.} \quad Xu Han$^{1,3}$\samethanks[1]{} \quad Yankai Lin$^5$ \quad Hexu Zhao$^{1,3}$\\ \textbf{Zhiyuan Liu}$^{1,2,3,6}$\thanks{ \ \ Corresponding authors.}
\quad \textbf{Peng Li}$^4$\thanks{\ \ Part of the work was done while Peng Li was working at Tencent.} \quad \textbf{Maosong Sun}$^{1,2,3,6\dagger}$ \quad \textbf{Jie Zhou}$^5$\\
  $^1$Department of Computer Science and Technology, Tsinghua University \\
  $^2$International Innovation Center of Tsinghua University,  \\
  $^3$Institute for Artificial Intelligence, Tsinghua University,  \\
  $^4$Institute for AI Industry Research (AIR), Tsinghua University\\
  $^5$Pattern Recognition Center, WeChat AI, Tencent Inc\\
  $^6$Beijing Academy of Artificial Intelligence\\
  \texttt{\{chenwz21,hanxu17\}@mails.tsinghua.edu.cn}\\
  \texttt{\{liuzy,sms\}@tsinghua.edu.cn}\\
  }

\begin{document}
\maketitle
\begin{abstract}
  Hyperbolic neural networks have shown great potential for modeling complex data. However, existing hyperbolic networks are not completely hyperbolic, as they encode features in the hyperbolic space yet formalize most of their operations in the tangent space (a Euclidean subspace) at the origin of the hyperbolic model. This hybrid method greatly limits the modeling ability of networks. In this paper, we propose a fully hyperbolic framework to build hyperbolic networks based on the Lorentz model by adapting the Lorentz transformations (including boost and rotation) to formalize essential operations of neural networks. Moreover, we also prove that linear transformation in tangent spaces used by existing hyperbolic networks is a relaxation of the Lorentz rotation and does not include the boost, implicitly limiting the capabilities of existing hyperbolic networks. The experimental results on four NLP tasks show that our method has better performance for building both shallow and deep networks. Our code is released to facilitate follow-up research\footnote{\url{https://github.com/chenweize1998/fully-hyperbolic-nn}}.
\end{abstract}

\section{Introduction}

Various recent efforts have explored hyperbolic neural networks to learn complex non-Euclidean data properties. 
\citet{nickel2017poincare,cvetkovski2016multidimensional,verbeek2014metric} learn hierarchical representations in a hyperbolic space and show that hyperbolic geometry can offer more flexibility than Euclidean geometry when modeling complex data structures. 
After that, \citet{ganea2018hyperbolic} and \citet{nickel2018learning} propose hyperbolic frameworks based on the \poincare ball model and the Lorentz model respectively\footnote{Both the \poincare ball model and the Lorentz model are typical geometric models in hyperbolic geometry.} to build hyperbolic networks, including hyperbolic feed-forward, hyperbolic multinomial logistic regression, etc. 

Encouraged by the successful formalization of essential operations in hyperbolic geometry for neural networks, various Euclidean neural networks are adapted into hyperbolic spaces. These efforts have covered a wide range of scenarios, from shallow neural networks like word embeddings~\cite{tifrea2018poincare,zhu2020hypertext}, network embeddings~\cite{chami2019hyperbolic,liu2019hyperbolic}, knowledge graph embeddings~\cite{balazevic2019multi,kolyvakis2019hyperkg} and attention module~\cite{gulcehre2018hyperbolic}, to deep neural networks like variational auto-encoders~\cite{mathieu2019continuous} and flow-based generative models~\cite{bose2020latent}. Existing hyperbolic neural networks equipped with low-dimensional hyperbolic feature spaces can obtain comparable or even better performance than high-dimensional Euclidean neural networks.

Although existing hyperbolic neural networks have achieved promising results, they are not fully hyperbolic. In practical terms, some operations in Euclidean neural networks that we usually use, such as matrix-vector multiplication, are difficult to be defined in hyperbolic spaces. Fortunately for each point in hyperbolic space, the tangent space at this point is a Euclidean subspace, all Euclidean neural operations can be easily adapted into this tangent space. Therefore, existing works~\cite{ganea2018hyperbolic,nickel2018learning} formalize most of the operations for hyperbolic neural networks in a hybrid way, by transforming features between hyperbolic spaces and tangent spaces via the logarithmic and exponential maps, and performing neural operations in tangent spaces. However, the logarithmic and exponential maps require a series of hyperbolic and inverse hyperbolic functions. The compositions of these functions are complicated and usually range to infinity, significantly weakening the stability of models. 

To avoid complicated transformations between hyperbolic spaces and tangent spaces, we propose a fully hyperbolic framework by formalizing operations for neural networks directly in hyperbolic spaces rather than tangent spaces. Inspired by the theory of special relativity, which uses Minkowski space (a Lorentz model) to measure the spacetime and formalizes the linear transformations in the spacetime as the Lorentz transformations, our hyperbolic framework selects the Lorentz model as our feature space. Based on the Lorentz model, we formalize operations via the relaxation of the Lorentz transformations to build hyperbolic neural networks, including linear layer, attention layer, etc. We also prove that performing linear transformation in the tangent space at the origin of hyperbolic spaces~\cite{ganea2018hyperbolic,nickel2018learning} is equivalent to performing a Lorentz rotation with relaxed restrictions, i.e., existing hyperbolic networks do not include the Lorentz boost, implicitly limiting their modeling capabilities. 

To verify our framework, we build fully hyperbolic neural networks for several representative scenarios, including knowledge graph embeddings, network embeddings, fine-grained entity typing, machine translation, and dependency tree probing. The experimental results show that our fully hyperbolic networks can outperform Euclidean baselines with fewer parameters. Compared with existing hyperbolic networks that rely on tangent spaces, our fully hyperbolic networks are faster, more stable, and achieve better or comparable results.

\section{Preliminaries}



Hyperbolic geometry is a non-Euclidean geometry with constant negative curvature $K$. Several hyperbolic geometric models have been applied in previous studies: the \poincare ball (\poincare disk) model~\cite{ganea2018hyperbolic}, the \poincare half-plane model~\cite{tifrea2018poincare}, the Klein model~\cite{gulcehre2018hyperbolic} and the Lorentz (Hyperboloid) model~\cite{nickel2018learning}. All these hyperbolic models are isometrically equivalent, i.e., any point in one of these models can be transformed to a point of others with distance-preserving transformations~\cite{ramsay1995introduction}. We select the Lorentz model as the framework cornerstone, considering the numerical stability and calculation simplicity of its exponential/logarithm maps and distance function. 

\subsection{The Lorentz Model}
\label{sec:bg-lorentz-model}

Formally, an $n$-dimensional Lorentz model is the Riemannian manifold $\lorentz = (\Lc^n, \riemanntensor)$. $K$ is the constant negative curvature. $\riemanntensor=\mathop{\mathgroup\symoperators diag}(-1,1,\cdots,1)$ is the Riemannian metric tensor. Each point in $\lorentz$ has the form $\mathbf{x} = \left[\begin{smallmatrix}
x_t\\
\mathbf{x}_s\\
\end{smallmatrix}\right], \mathbf{x}\in\Real^{n+1}, x_t \in \Real, \mathbf{x}_s \in \Real^n$. $\Lc^n$ is a point set satisfying 
\begin{align*}
\Lc^n \coloneqq \{\mathbf{x}\in &\Real^{n+1}  \mid \linner{x}{x}=\frac{1}{K}, x_t >0\},\\
\linner{x}{y}&\coloneqq-x_t y_t+\mathbf{x}_s^\intercal\mathbf{y}_s \\
&=\mathbf{x}^\intercal \mathop{\mathgroup\symoperators diag}(-1,1,\cdots,1) \mathbf{y},
\end{align*}
where $\linner{x}{y}$ is the Lorentzian inner product, $\Lc^n$ is the upper sheet of hyperboloid (hyper-surface) in an $(n+1)$-dimensional Minkowski space with the origin $(\sqrt{-1/K}, 0, \cdots, 0)$. For simplicity, we denote a point $\mathbf{x}$ in the Lorentz model as $\mathbf{x} \in \lorentz$ in the latter sections. 

The special relativity gives physical interpretation to the Lorentz model by connecting the last $n$ elements $\mathbf{x}_s$ to \textit{space} and the $0$-th element $x_t$ to \textit{time}. We follow this setting to denote the $0$-th dimension of the Lorentz model as \textit{time axis}, and the last $n$ dimensions as \textit{spatial axes}.

\paragraph{Tangent Space}
\label{sec:bg-lorentz-model-tangent} 

Given $\mathbf{x}\in\lorentz$, the orthogonal space of $\lorentz$ at $\mathbf{x}$ with respect to the Lorentzian inner product is the tangent space at $\mathbf{x}$, and is formally written as
\begin{align*}
\tangent{x}\coloneqq\{\mathbf{y}\in \Real^{n+1} \mid \linner{y}{x}=0\}.
\end{align*}
Note that $\tangent{x}$ is a Euclidean subspace of $\Real^{n+1}$. Particularly, we denote the tangent space at the origin as $\tangent{0}$. 

\paragraph{Logarithmic and Exponential Maps}
\label{sec:bg-lorentz-model-exp}

\begin{figure*}[t]
\centering
\subfloat[Linear layer formalized in tangent space]{
\includegraphics[width=0.32\linewidth]{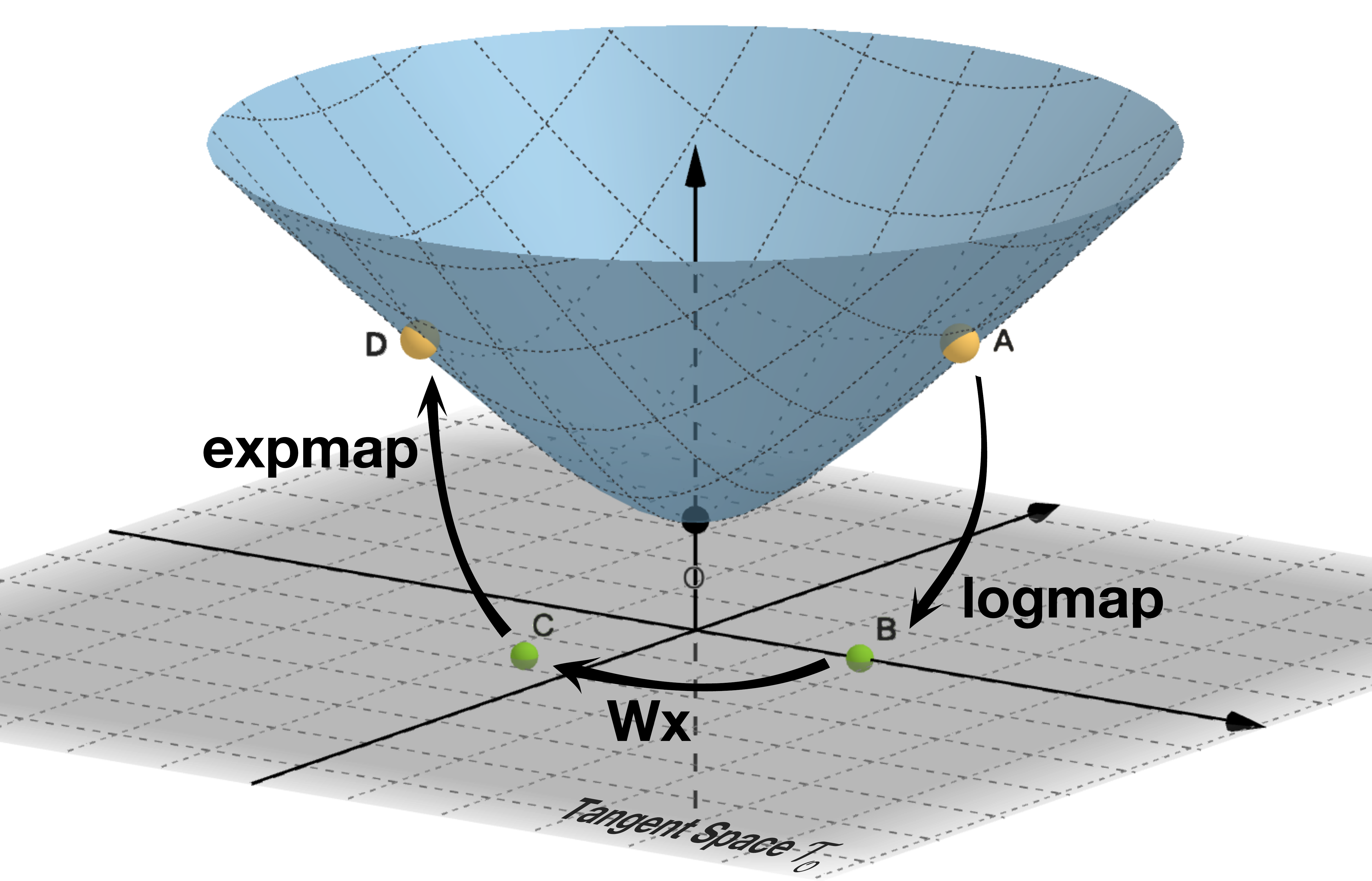}
\label{fig:logexp}
}
\subfloat[Lorentz boost]{
\includegraphics[width=0.21\linewidth]{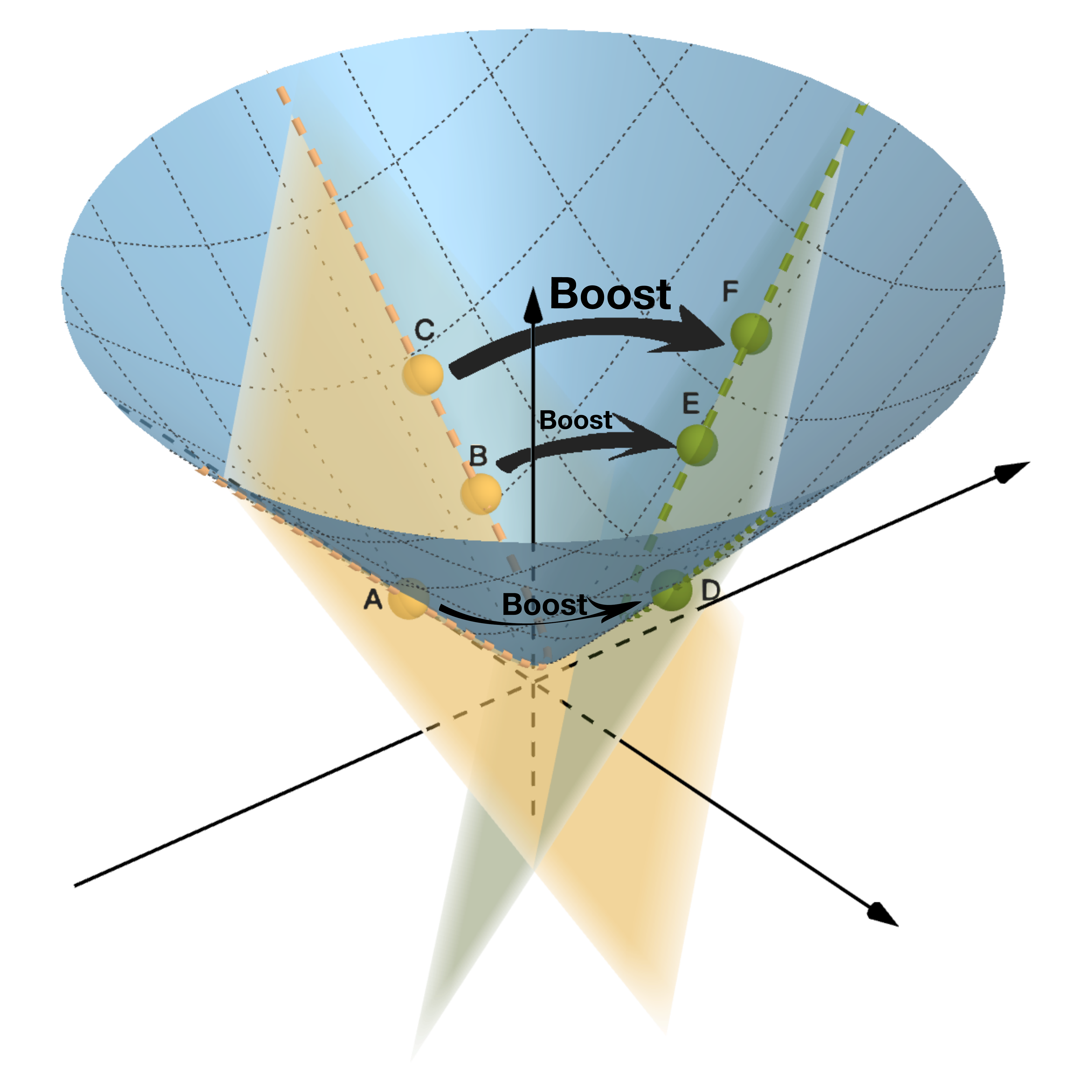}
\label{fig:lorentz-boost}
}
\subfloat[Lorentz rotation]{
\includegraphics[width=0.21\linewidth]{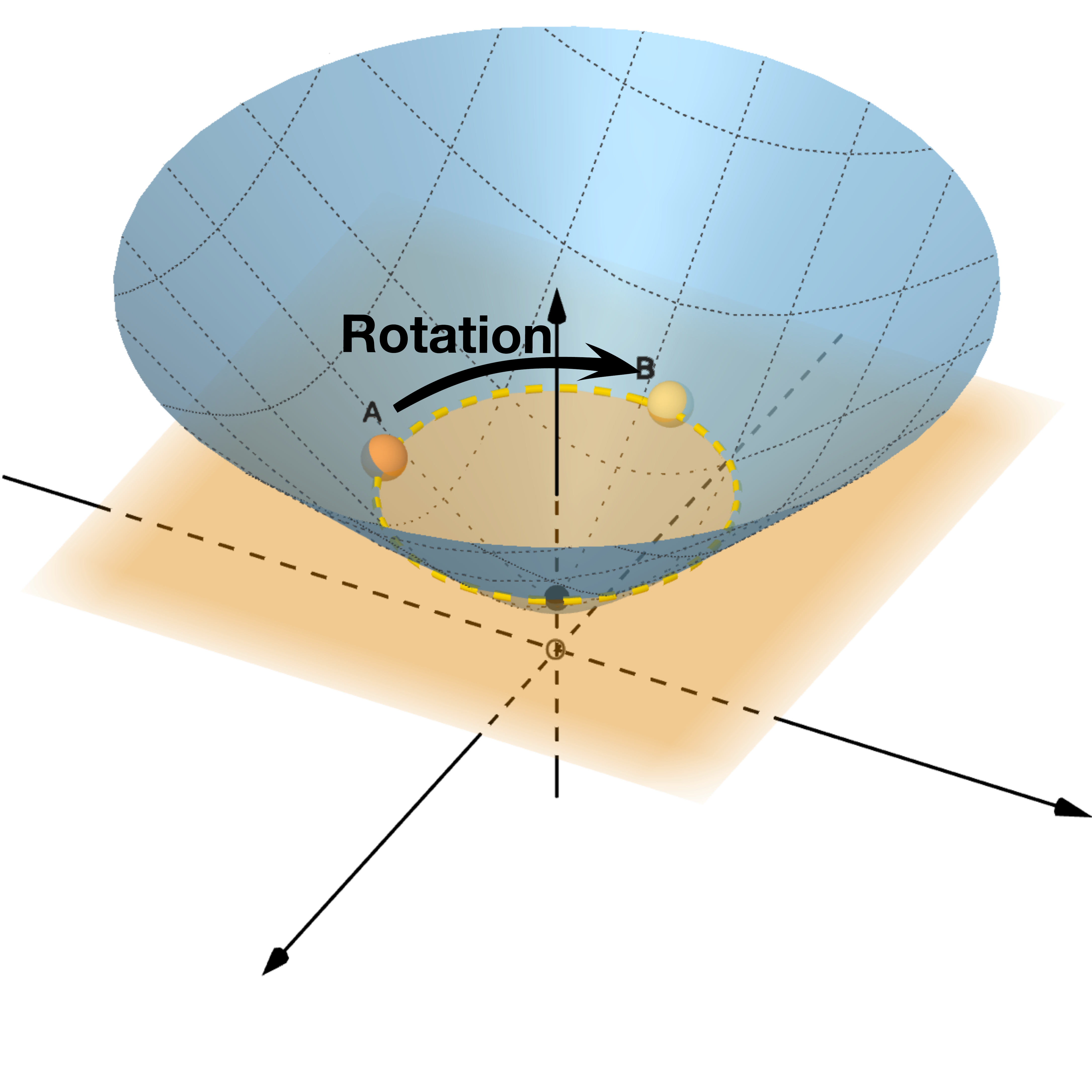}
\label{fig:lorentz-rotation}
}
\subfloat[Pseudo-rotation]{
\includegraphics[width=0.21\linewidth]{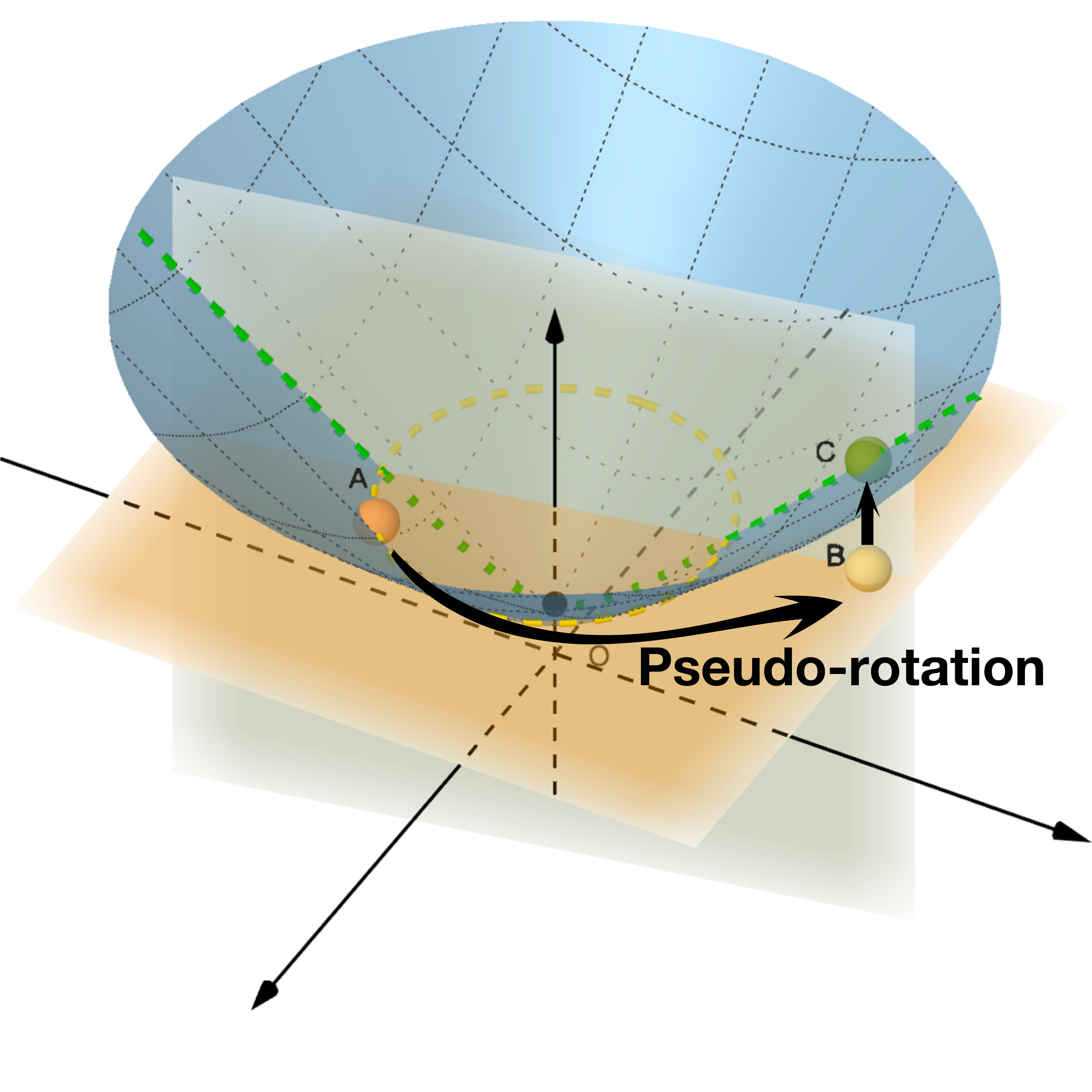}
\label{fig:pseudo-rotation}
}
\caption{Illustration of a hyperbolic linear layer based on the logarithmic and exponential maps as well as different transformations in the Lorentz model. In \cref{fig:logexp}, $A$ is mapped to $B$ in the tangent space at the origin $\tangent{0}$ through the logarithmic map. A Euclidean linear transformation is performed to obtain $C$. Finally, $C$ is mapped back to the hyperbolic space through the exponential map. \cref{fig:lorentz-boost,fig:lorentz-rotation} are the visualization of the Lorentz boost and rotation, where points on the intersection of a plane and the hyperboloid are still coplanar after the Lorentz boost. \cref{fig:pseudo-rotation} is pseudo-rotation in \cref{sec:relation-tangent}, where a point is first transformed and then projected onto the hyperboloid.}
\label{fig:lorentz-transformation}
\end{figure*}

As shown in Figure~\ref{fig:logexp}, the logarithmic and exponential maps specifies the mapping of points between the hyperbolic space $\lorentz$ and the Euclidean subspace $\tangent{x}$.

The exponential map $\exp_{\mathbf{x}}^K(\mathbf{z}):\tangent{x}\rightarrow\lorentz$ can map any tangent vector $\mathbf{z} \in \tangent{x}$ to $\lorentz$ by moving along the geodesic $\gamma$ satisfying $\gamma(0)=\mathbf{x}$ and $\gamma'(0)=\mathbf{z}$. Specifically, the exponential map can be written as
\begin{align*}
\expmap{x}{z} &= \cosh(\alpha)\mathbf{x} + \sinh(\alpha)\frac{\mathbf{z}}{\alpha},\\ 
\alpha&=\sqrt{-K}\lVert\mathbf{z}\rVert_\Lc,\\ \lVert\mathbf{z}\rVert_\Lc &= \sqrt{\linner{z}{z}}.
\end{align*}

The logarithmic map $\logmap{\mathbf{x}}{\mathbf{y}}:\lorentz \rightarrow \tangent{x}$ plays an opposite role to map $\mathbf{y} \in \lorentz$ to $\tangent{x}$. and it can be written as
\begin{align*}
\logmap{\mathbf{x}}{\mathbf{y}} &= \frac{\cosh^{-1}(\beta)}{\sqrt{\beta^2-1}}(\mathbf{y}-\beta \mathbf{x}),\\ \beta&=K\linner{x}{y}.
\end{align*}



\subsection{The Lorentz Transformations}
\label{sec:the-lorentz-transformations}

In the special relativity, the Lorentz transformations are a family of linear transformations from a coordinate frame in spacetime to another frame moving at a constant velocity relative to the former. Any Lorentz transformation can be decomposed into a combination of a Lorentz boost and a Lorentz rotation by polar decomposition~\cite{moretti2002interplay}.
\begin{Definition}[Lorentz Boost]
Lorentz boost describes relative motion with constant velocity and without rotation of the spatial coordinate axes. Given a velocity $\mathbf{v} \in \Real^n$ (ratio to the speed of light), $\lVert \mathbf{v} \rVert < 1$ and $\gamma=\frac{1}{\sqrt{1-\lVert\mathbf{v}\rVert^2}}$, the Lorentz boost matrices are given by $\mathbf{B} = 
\left[\begin{array}{cccc}
\gamma & -\gamma \mathbf{v}^\intercal \\
-\gamma \mathbf{v}& \mathbf{I}+\frac{\gamma^2}{1+\gamma}\mathbf{v}\mathbf{v}^\intercal \\
\end{array}\right]$.
\end{Definition}

\begin{Definition}[Lorentz Rotation]
Lorentz rotation is the rotation of the spatial coordinates. The Lorentz rotation matrices are given by $\mathbf{R} = \left[\begin{array}{cccc}
1 & \mathbf{0}^\intercal\\
\mathbf{0} & \mathbf{\tilde{R}} \\
\end{array}\right]$,
where $\mathbf{\tilde{R}}^\intercal\mathbf{\tilde{R}} = \mathbf{I}$ and $\det(\mathbf{\tilde{R}}) = 1$, i.e., $\mathbf{\tilde{R}} \in \textbf{SO}(n)$ is a special orthogonal matrix.
\end{Definition}

Both the Lorentz boost
and the Lorentz rotation are the linear transformations directly defined in the Lorentz model, i.e., $\forall \mathbf{x}\in \lorentz, \mathbf{B} \mathbf{x} \in \lorentz$ and $\mathbf{R} \mathbf{x} \in \lorentz$. Hence, we build fully hyperbolic neural networks on the basis of these two types of transformations in this paper.

\section{Fully Hyperbolic Neural Networks}
\label{sec:components}



\subsection{Fully Hyperbolic Linear Layer}
\label{sec:components-linear}


We first introduce our hyperbolic linear layer in the Lorentz model, considering it is the most essential block for neural networks. Although the Lorentz transformations in \cref{sec:the-lorentz-transformations} are linear transformations in the Lorentz model, they cannot be directly used for neural networks. On the one hand, the Lorentz transformations transform coordinate frames without changing the number of dimensions. On the other hand, complicated restrictions of the Lorentz transformations (e.g., special orthogonal matrices for the Lorentz rotation) make computation and optimization problematic. Although the restrictions offer nice properties such as spacetime interval invariant to Lorentz transformation, they are unwanted in neural networks.

A Lorentz linear layer matrix should minimize the loss while subject to $\mathbf{M}\in\Real^{(m+1)\times (n+1)}$, $\forall \mathbf{x} \in \mathbb{L}^n, \mathbf{M} \mathbf{x} \in \mathbb{L}^m$. It is a constrained optimization difficult to solve. We instead re-formalize our lorentz linear layer to learn a matrix $\mathbf{M} = \left[\begin{smallmatrix}
\mathbf{v}^\intercal\\
\mathbf{W}\\
\end{smallmatrix}\right], \mathbf{v} \in \Real^{n + 1}, \mathbf{W} \in \Real^{m \times (n + 1)}$ satisfying $\forall \mathbf{x} \in \mathbb{L}^n, f_{\mathbf{x}}(\mathbf{M}) \mathbf{x} \in \mathbb{L}^m$, where $f_{\mathbf{x}}: \Real^{(m+1) \times (n+1)} \rightarrow \Real^{(m+1) \times (n+1)}$ should be a function that maps any matrix to a suitable one for the hyperbolic linear layer. Specifically, $\forall \mathbf{x} \in \lorentz, \mathbf{M}\in\Real^{(m+1)\times(n+1)}$,  $f_{\mathbf{x}}(\mathbf{M})$ is given as
\begin{equation}
\label{eq:linear-matrix}
\small
f_{\mathbf{x}}(\mathbf{M}) = f_{\mathbf{x}}(\left[\begin{array}{c}
\mathbf{v}^\intercal\\
\mathbf{W}\\
\end{array}\right]) = \left[\begin{array}{c}
\frac{\sqrt{\lVert \mathbf{W}\mathbf{x} \rVert^2 - 1/K}}{\mathbf{v}^\intercal \mathbf{x}} \mathbf{v}^\intercal\\
\mathbf{W}\\
\end{array}\right],
\end{equation}

\begin{theorem} 
\label{theorem0}
$\forall \mathbf{x} \in \lorentz, \forall \mathbf{M} \in \Real^{(m+1)\times(n+1)}$, we have $f_{\mathbf{x}}(\mathbf{M})\mathbf{x} \in \mathbb{L}^m_K$.
\end{theorem}
\begin{Proof}
One can easily verify that
$\forall \mathbf{x}\in\lorentz$, we have $\langle f_{\mathbf{x}}(\mathbf{M})\mathbf{x}, f_{\mathbf{x}}(\mathbf{M})\mathbf{x}\rangle_\mathcal{L} = 1/K$, thus $f_\mathbf{x}(\mathbf{M})\mathbf{x}\in\mathbb{L}^m_K$.
\qed
\end{Proof} 


\paragraph{Relation to the Lorentz Transformations}

In this part, we show that the set of matrices $\{f_\mathbf{x}(\mathbf{M})\}$ defined in Eq.\eqref{eq:linear-matrix} contains all Lorentz rotation and boost matrices.

\begin{Lemma} 
\label{lemma1}
In the $n$-dimensional Lorentz model $\lorentz$, we denote the set of all Lorentz boost matrices as $\mathcal{B}$
, the set of all Lorentz rotation matrices as $\mathcal{R}$
. Given $\mathbf{x} \in \lorentz$, we denote the set of $f_{\mathbf{x}}(\mathbf{M})$ at $\mathbf{x}$ without changing the number of space dimension as $\mathcal{M}_{\mathbf{x}} =\{ f_{\mathbf{x}}(\mathbf{M}) \mid \mathbf{M} \in \Real^{(n+1) \times (n+1)} \}$. $\forall \mathbf{x} \in \lorentz$, we have $ \mathcal{B} \subseteq  \mathcal{M}_{\mathbf{x}}$ and $ \mathcal{R} \subseteq  \mathcal{M}_{\mathbf{x}}$.
\end{Lemma}
\begin{Proof} 
\label{proof1}
We first prove $\mathcal{M}_\mathbf{x}$ covers all valid transformations.

Considering $\mathcal{A}=\{\mathbf{A}\in\Real^{(n+1)\times(n+1)}\mid\forall \mathbf{x}\in\lorentz: \langle \mathbf{A}\mathbf{x}, \mathbf{A}\mathbf{x}\rangle_\mathcal{L}=\frac{1}{K}, (\mathbf{Ax})_0>0\}$ is the set of all valid transformation matrices in the Lorentz model. Then $\forall\mathbf{A}=\left[\begin{smallmatrix}
\mathbf{v}^\intercal_{A}\\
\mathbf{W}_{A}\\
\end{smallmatrix}\right] \in \mathcal{A}$, $\mathbf{v}_A\in\Real^{n+1}, \mathbf{W}_A\in\Real^{n\times(n+1)}$, $\exists\mathbf{x}\in\Real^{n+1}\colon \mathbf{v}^\intercal \mathbf{x} > 0$ and $\lVert \mathbf{W}_{A} \mathbf{x} \rVert^2 - {(\mathbf{v}^\intercal_A \mathbf{x})}^2  = \frac{1}{K}$. Furthermore, $\forall \mathbf{A}\in\mathcal{A}$, we have
\begin{align*}
f_{\mathbf{x}}(\mathbf{A}) = f_{\mathbf{x}}(\left[\begin{smallmatrix}
\mathbf{v}^\intercal_{A}\\
\mathbf{W}_{A}\\
\end{smallmatrix}\right]) = \left[\begin{smallmatrix}
\frac{\sqrt{\lVert \mathbf{W}_{A}\mathbf{x} \rVert^2 -1/K}}{\mathbf{v}^\intercal_{A} \mathbf{x}} \mathbf{v}_{A}^\intercal\\
\mathbf{W}_{A}\\
\end{smallmatrix}\right] = \mathbf{A}.
\end{align*} 
Hence, we can see that $\mathcal{A}\subseteq \mathcal{M}_\mathbf{x}$. Since $\mathcal{B}\subseteq\mathcal{A}$ and $\mathcal{R}\subseteq\mathcal{A}$, therefore $\mathcal{B}\subseteq\mathcal{M}_\mathbf{x}$ and $\mathcal{R}\subseteq\mathcal{M}_\mathbf{x}$.
\qed
\end{Proof} 
According to \cref{theorem0,lemma1}, both Lorentz boost and rotation can be covered by our linear layer.

\paragraph{Relation to the Linear Layer Formalized in the Tangent Space}
\label{sec:relation-tangent}

In this part, we show that the conventional hyperbolic linear layer formalized in the tangent space at the origin~\cite{ganea2018hyperbolic,nickel2018learning} can be considered as a Lorentz transformation with only a special rotation but no boost. \cref{fig:logexp} visualizes the conventional hyperbolic linear layer.

As shown in Figure~\ref{fig:pseudo-rotation}, we consider a special setting ``\textit{pseudo-rotation}" of our hyperbolic linear layer. Formally, at the point $\mathbf{x} \in \lorentz$, all pseudo-rotation matrices make up the set $\mathcal{P}_{\mathbf{x}} =\big\{ f_{\mathbf{x}}( \left[\begin{smallmatrix}
  w       & \mathbf{0}^\intercal\\
\mathbf{0} & \mathbf{W}
\end{smallmatrix}\right]) \mathrel{\big|} w \in \Real, \mathbf{W} \in \Real^{n \times n} \big\}$. As we no longer require the submatrix $\mathbf{W}$ to be a special orthogonal matrix, this setting is a relaxation of the Lorentz rotation.

Formally, given $\mathbf{x} \in \lorentz$, the conventional hyperbolic linear layer relies on the logarithmic map to map the point into the tangent space at the origin, a matrix to perform linear transformation in the tangent space, and the exponential map to map the final result back to $\lorentz$ \footnote{Note that Mobius matrix-vector multiplication defined in~\citet{ganea2018hyperbolic} also follows this process}. The whole process \footnote{The $0$-th dimension of any point in the tangent space at the origin is $0$, therefore the linear matrix has the form $\mathop{\mathgroup\symoperators diag}(*, \mathbf{W})$, where $*$ can be arbitrary number.} is
\begin{equation}
  \small
\begin{aligned}
\exp_{\mathbf{0}}(
\left[\begin{smallmatrix}
*    & \mathbf{0}^\intercal \\
\mathbf{0} & \mathbf{W}
\end{smallmatrix}\right]
\log_{\mathbf{0}}(
[\begin{smallmatrix}
x_t\\
\mathbf{x}_s
\end{smallmatrix}]
))=\left[\begin{smallmatrix}
\frac{\cosh(\beta)}{\sqrt{-K}x_t}    & \mathbf{0}^\intercal \\
\mathbf{0} & \frac{\sinh(\beta)\mathbf{W}}{\sqrt{-K}\lVert \mathbf{W}\mathbf{x}_s \rVert }
\end{smallmatrix}\right]
[\begin{smallmatrix}
x_t\\
\mathbf{x}_s
\end{smallmatrix}],\\
\end{aligned}
\label{eq:logexp}
\end{equation}
where $\beta = \frac{\sqrt{-K}\cosh^{-1}(\sqrt{-K}x_t)}{\sqrt{-Kx_t^2-1}} \lVert \mathbf{W}\mathbf{x}_s \rVert$.

\begin{Lemma} 
\label{lemma2}
$\forall \mathbf{x} \in \lorentz$ , we define the set of the outcomes of Eq.\eqref{eq:logexp} as 
\begin{align*}
\mathcal{H}_{\mathbf{x}} =\left\{\left[\begin{smallmatrix}
\frac{\cosh(\beta)}{\sqrt{-K}x_t}    & \mathbf{0}^\intercal \\
\mathbf{0} & \frac{\sinh(\beta)}{\sqrt{-K}\lVert \mathbf{W}\mathbf{x}_s \rVert }\mathbf{W}
\end{smallmatrix}\right]\mathrel{\bigg|} \mathbf{W} \in \Real^{n\times n}
\right\},
\end{align*} 
then we have $\mathcal{H}_{\mathbf{x}} \subseteq  \mathcal{P}_{\mathbf{x}}$ and $\mathcal{H}_{\mathbf{x}} \cap \mathcal{B} = \{\mathbf{I}\}$.
\end{Lemma}

\begin{Proof}
$\forall\mathbf{x}\in\lorentz,\forall\mathbf{H} \in \mathcal{H}_{\mathbf{x}}$, $\mathbf{H}$ has the form $\left[\begin{smallmatrix}
w & \mathbf{0}^\intercal \\
\mathbf{0} & \mathbf{W}
\end{smallmatrix}\right]$, satisfying $\lVert \mathbf{W} \mathbf{x}_s \rVert^2 - {(w x_t)}^2  = \frac{1}{K}$ and $w x_t>0$. We can verify that
\begin{align*}
f_{\mathbf{x}}(\mathbf{H}) = f_{\mathbf{x}}(\left[\begin{smallmatrix}
w & \mathbf{0}^\intercal \\
\mathbf{0} & \mathbf{W}
\end{smallmatrix}\right]) = \left[\begin{smallmatrix}
\frac{\sqrt{\lVert \mathbf{W} \mathbf{x}_{s} \rVert^2 - 1/K}}{w  x_t} w & \mathbf{0}^\intercal\\
\mathbf{0} & \mathbf{W}\\
\end{smallmatrix}\right] = \mathbf{H}.
\end{align*} 
Hence, $\forall \mathbf{x} \in \lorentz$, $\forall \mathbf{H} \in \mathcal{H}_{\mathbf{x}}$, we have $\mathbf{H} = f_{\mathbf{x}}(\mathbf{H}) \in \mathcal{P}_{\mathbf{x}}$, and thus $\mathcal{H}_{\mathbf{x}} \subseteq \mathcal{P}_{\mathbf{x}}$. 
\qed
\end{Proof} 

To prove $\mathcal{H}_{\mathbf{x}} \cap \mathcal{B} = {\mathbf{I}}$ is trivial, we do not elaborate here. Therefore, a conventional hyperbolic linear layer can be considered as a special rotation where the time axis is changed according to the space axes to ensure that the output is still in the Lorentz model. Our linear layer is not only fully hyperbolic but also equipped with boost operations to be more expressive. Moreover, without using the complicated logarithmic and exponential maps, our linear layer has better efficiency and stability. 

\paragraph{A More General Formula}
Here, we give a more general formula\footnote{Note that this general formula is no longer fully hyperbolic. It is a relaxation in implementation, while the input and output are still guaranteed to lie in the Lorentz model.} of our hyperbolic linear layer based on $f_{\mathbf{x}}(\left[\begin{smallmatrix}
\mathbf{v}^\intercal\\
\mathbf{W}\\
\end{smallmatrix}\right])\mathbf{x}$, by adding activation, dropout, bias and normalization,
\begin{equation}
\label{eq:hlinear}
\mathbf{y} = \texttt{HL}(\mathbf{x}) = \left[\begin{smallmatrix}
\sqrt{\lVert \phi(\mathbf{Wx, v}) \rVert^2 - 1/K}  \\
\phi(\mathbf{W}\mathbf{x}, \mathbf{v})
\end{smallmatrix}\right],
\end{equation}
where $\mathbf{x} \in \lorentz$, $\mathbf{v} \in \Real^{n+1}$, $\mathbf{W} \in \Real^{m\times(n+1)}$, and $\phi$ is an operation function: for the dropout, the function is $\phi(\mathbf{Wx, v}) = \mathbf{W}\texttt{dropout}(\mathbf{x})$; for the activation and normalization $\phi(\mathbf{Wx, v}) = \frac{\lambda \sigma(\mathbf{v}^\intercal\mathbf{x}+b')}{\lVert \mathbf{W}h(\mathbf{x})+\mathbf{b} \rVert}(\mathbf{W}h(\mathbf{x)+b})$, where $\sigma$ is the sigmoid function, $\mathbf{b}$ and $b'$ are bias terms, $\lambda > 0$ controls the scaling range, $h$ is the activation function. We elaborate $\phi(\cdot)$ we use in practice in the appendix.

\subsection{Fully Hyperbolic Attention Layer}
\label{sec:components-attention}

Attention layers are also important for building networks, especially for the networks of Transformer family~\cite{vaswani2017attention}. We propose an attention module in the Lorentz model. Specifically, we consider the weighted aggregation of a point set $\mathcal{P} = \{\mathbf{x}_1,\ldots, \mathbf{x}_{|\mathcal{P}|}\}$ as calculating the centroid, whose expected (squared) distance to $\mathcal{P}$ is minimum, i.e. $\argmin_{\bm{\mu}\in\lorentz}\sum_{i=1}^{|\mathcal{P}|}\nu_i d_\mathcal{L}^2(\mathbf{x}_i, \bm{\mu})$, where $\nu_i$ is the weight of the $i$-th point. \citet{law2019lorentzian} prove that, with squared Lorentzian distance defined as $d^2_\Lc(\mathbf{a}, \mathbf{b})=2/K-2\linner{a}{b}$, the centroid \textit{w.r.t.} the squared Lorentzian distance is given as 
\begin{equation}
\label{eq:mass-center}
\begin{aligned}
  \bm{\mu}&=\texttt{Centroid}\big(\{\nu_1,\ldots,\nu_{|\mathcal{P}|}\}, \{\mathbf{x}_1,\ldots,\mathbf{x}_{|\mathcal{P}|}\}\big)\\
  &=\frac{\sum_{j=1}^{|\mathcal{P}|} \nu_{j}\mathbf{x}_j}{\sqrt{-K} \big| \lnorm{\sum_{i=1}^{ |\mathcal{P}|}\nu_{i}\mathbf{x}_i}\big|}.
\end{aligned}
\end{equation}

Given the query set $\mathcal{Q} = \{\mathbf{q}_1,\ldots, \mathbf{q}_{|\mathcal{Q}|} \}$, key set  $\mathcal{K} = \{\mathbf{k}_1, \ldots, \mathbf{k}_{|\mathcal{K}|} \}$, and value set $\mathcal{V} = \{\mathbf{v}_1, \ldots, \mathbf{v}_{|\mathcal{V}|} \}$, where $|\mathcal{K}| = |\mathcal{V}|$, we exploit the squared Lorentzian distance between points to calculate weights. The attention is defined as $\texttt{ATT}(\mathcal{Q}, \mathcal{K}, \mathcal{V}) = \{\bm{\mu}_1, \ldots, \bm{\mu}_{|\mathcal{Q}|}\}$ and given by:
\begin{equation}
  \begin{aligned}
    \bm{\mu}_i&=\frac{\sum_{j=1}^{|\mathcal{K}|} \nu_{ij}\mathbf{v}_{j}}{\sqrt{-K} \big | \lnorm{\sum_{k=1}^{|\mathcal{K} |}\nu_{ik}\mathbf{v}_k} \big|},\\
    \nu_{ij} &= \frac{\exp (\frac{-d_{\Lc}^2(\mathbf{q}_i, \mathbf{k}_j)}{\sqrt{n}})}{\sum_{k=1}^{|\mathcal{K}|} \exp (\frac{-d_{\Lc}^2(\mathbf{q}_i, \mathbf{k}_k)}{\sqrt{n}})},
  \end{aligned}
\end{equation}
where $n$ is the dimension of points. Furthermore, multi-headed attention is defined as $\texttt{MHATT} (\mathcal{Q}, \mathcal{K}, \mathcal{V}) = \{\bm{\mu}_1, \ldots, \bm{\mu}_{|\mathcal{Q}|}\}$, and $\bm{\mu}_i$ is 
\begin{equation}
  \begin{aligned}
    \bm{\mu}_i &= \texttt{HL}([\bm{\mu}_i^1|\ldots|\bm{\mu}_i^H]), \\
    \{\bm{\mu}_1^i,\bm{\mu}_2^i,\ldots\} &= 
      \texttt{ATT}^{i}(\texttt{HL}_{\mathcal{Q}}^{i}(\mathcal{Q}), \texttt{HL}_{\mathcal{K}}^{i}(\mathcal{K}), \texttt{HL}_{\mathcal{V}}^{i}(\mathcal{V})),
  \end{aligned}
\end{equation}
where $H$ is the head number, $[\cdot|\ldots|\cdot]$ is the concatenation of multiple vectors, $\texttt{ATT}^{i}(\cdot,\cdot,\cdot)$ is the $i$-th head attention, and $\texttt{HL}_{\mathcal{Q}}^{i}(\cdot)$, $\texttt{HL}_{\mathcal{K}}^{i}(\cdot)$, $\texttt{HL}_{\mathcal{V}}^{i}(\cdot)$ are the hyperbolic linear layers of the $i$-th head attention. 

Other intuitive choices for the aggregation in the Lorentz attention module include Fréchet mean~\cite{https://doi.org/10.1002/cpa.3160300502} and Einstein midpoint~\cite{ungar2005analytic}. The Fréchet mean is the classical generalization of Euclidean mean. However, it offers no closed-form solution. Solving the Fréchet mean currently requires iterative computation~\cite{lou2020differentiating,gu2018learning}, which significantly slows down the training and inference, making it impossible to generalize to deep and large model \footnote{400 times slower than using Lorentz centroid in our experiment, and no improvement in performance was observed}. On the contrary, Lorentz centroid is fast to compute and can be seen as Frechet mean in pseudo-hyperbolic space~\cite{law2019lorentzian}. The computation of the Einstein midpoint requires transformation between Lorentz model and Klein model, bringing in numerical instability. The Lorentz centroid we use minimizes the sum of squared distance in the Lorentz model, while the Einstein midpoint does not possess such property in theory. Also, whether the Einstein midpoint in the Klein model has its geometric interpretation in the Lorentz model remains to be investigated, and it is beyond the scope of our paper. Therefore, we adopt the Lorentz centroid in our Lorentz attention.

\subsection{Fully Hyperbolic Residual Layer and Position Encoding Layer}
\label{sec:components-residual-embedding}
\paragraph{Lorentz Residual}
The residual layer is crucial for building deep neural networks. Since there is no well-defined vector addition in the Lorentz model, we assume that each residual layer is preceded by a computational block whose last layer is a Lorentz linear layer, and do the residual-like operation within the preceding Lorentz linear layer of the block as a compromise. Given the input $\mathbf{x}$ of the computational block and the output $\mathbf{o} = f(\mathbf{x})$ before the last Lorentz linear layer of the block, we take $\mathbf{x}$ as the bias of the Lorentz linear layer. Concretely, the final output of the block is
\begin{equation}
\label{eq:lresidual}
  \begin{aligned}
      \mathbf{y}&=\left[\begin{smallmatrix}
\sqrt{\lVert \phi(\mathbf{W}\mathbf{o}, \mathbf{v}, \mathbf{x}) \rVert^2 - 1/K}  \\
\phi(\mathbf{W}\mathbf{o}, \mathbf{v}, \mathbf{x})
\end{smallmatrix}\right], \\
\phi(\mathbf{W}\mathbf{o}, \mathbf{v}, \mathbf{x})&=\frac{\lambda \sigma(\mathbf{v}^\intercal \mathbf{o})}{\lVert \mathbf{W}h(\mathbf{o})+\mathbf{x}_{s} \rVert}(\mathbf{W}h(\mathbf{o})+\mathbf{x}_{s}),
  \end{aligned}
\end{equation}
where the symbols have the same meaning as those in Eq.\eqref{eq:hlinear}.

\paragraph{Lorentz Position Encoding}
Some neural networks require positional encoding for their embedding layers, especially those models for NLP tasks. Previous works generally incorporate positional information by adding position embeddings to word embeddings. Given a word embedding $\mathbf{x}$ and its corresponding learnable position embedding $\mathbf{p}$, we add a Lorentz linear layer to transform the word embedding $\mathbf{x}$, by taking the position embedding $\mathbf{p}$ as the bias. The overall process is the same as Eq.\eqref{eq:lresidual}. Note that the transforming matrix in the Lorentz linear layer is shared across positions. This modification gives us one more $d\times d$ matrix than the Euclidean Transformer. The increase in the number of parameters is negligible compared to the huge parameters of the whole model.

\bgroup
\setlength{\tabcolsep}{5pt}
\def\arraystretch{0.9}
\begin{table*}[t]
\small
\begin{center}
  \begin{tabular}{l*{10}{r}}
    \toprule
    & \multicolumn{5}{c}{\textbf{WN18RR}} & \multicolumn{5}{c}{\textbf{FB15k-237}} \\
    \cmidrule(lr){2-6}\cmidrule(lr){7-11}
    \textbf{Model} & \textbf{\#Dims} & \textbf{MRR} & \textbf{H@10} & \textbf{H@3} & \textbf{H@1} & \textbf{\#Dims} & \textbf{MRR} & \textbf{H@10} & \textbf{H@3} & \textbf{H@1}\\
    \midrule
    \textsc{TransE}~\cite{bordes2013translating} & 180 & 22.7 & 50.6 & 38.6 & 3.5 & 200 & 28.0 & 48.0 & 32.1 & 17.7 \\
    \textsc{DistMult}~\cite{yang2014embedding} & 270 & 41.5 & 48.5 & 43.0 & 38.1 & 200 & 19.3 & 35.3 & 20.8 & 11.5\\
    \textsc{ComplEx}~\cite{trouillon2017knowledge} & 230 & 43.2 & 50.0 & 45.2 & 39.6 & 200 & 25.7 & 44.3 & 29.3 & 16.5 \\
    \textsc{ConvE}~\cite{dettmers2018convolutional} & 120 & 43.5 & 50.0 & 44.6 & 40.1 & 200 & 30.4 & 49.0 & 33.5 & 21.3 \\
    \textsc{RotatE}~\cite{sun2019rotate} & 1,000 & 47.3 & 55.3 & 48.8 & 43.2 & 1,024 & 30.1 & 48.5 & 33.1 & 21.0 \\
    \textsc{TuckER}~\cite{balazevic2019tucker} & 200 & 46.1 & 53.5 & 47.8 & 42.3 & 200 & 34.7 & 53.3 & 38.4 & 25.4 \\
     \midrule
    \textsc{MuRP}~\cite{balazevic2019multi} & 32 & 46.5 & 54.4 & 48.4 & 42.0 & 32 & 32.3 & 50.1 & 35.3 & 23.5\\
    \textsc{RotH}~\cite{chami2020low} & 32 & 47.2 & \underline{55.3} & 49.0 & 42.8 & 32 & 31.4 & 49.7 & 34.6 & 22.3\\
    \textsc{AttH}~\cite{chami2020low} & 32 & 46.6 & 55.1 & 48.4 & 41.9 & 32 & 32.4 & 50.1 & 35.4 & 23.6\\
    \textsc{HyboNet} & 32 & \underline{48.9} & \underline{55.3} & \underline{50.3} & \underline{45.5} & 32 & \underline{33.4} & \underline{51.6} & \underline{36.5} & \underline{24.4} \\\midrule
    \textsc{MuRP}~\cite{balazevic2019multi} & $\beta$ & 48.1 & 56.6 & 49.5 & 44.0 & $\beta$ & 33.5 & 51.8 & 36.7 & 24.3\\
    \textsc{RotH}~\cite{chami2020low} & $\beta$ & 49.6 & \underline{\textbf{58.6}} & 51.4 & 44.9 & $\beta$ & 34.4 & 53.5 & 38.0 & 24.6\\
    \textsc{AttH}~\cite{chami2020low} & $\beta$ & 48.6 & 57.3 & 49.9 & 44.3 & $\beta$ & 34.8 & \underline{\textbf{54.0}} & 38.4 & 25.2\\
    \textsc{HyboNet} & $\beta$ & \underline{\textbf{51.3}} & 56.9 & \underline{\textbf{52.7}} & \underline{\textbf{48.2}} & $\beta$ & \underline{\textbf{35.2}} & 52.9 & \underline{\textbf{38.7}} & \underline{\textbf{26.3}} \\
   \bottomrule
  \end{tabular}
  \label{tab:exp-graph-link-prediction}
\end{center}
\caption{Link prediction results (\%) on WN18RR and FB15k-237 in the filtered setting. $\beta\in\{200, 400, 500\}$ and we report the best result. The first group of models are Euclidean models, the second and third groups are hyperbolic models with different dimensions. Following \citet{balazevic2019multi}, RotatE results are reported without their self-adversarial negative sampling for fair comparison. Best results are in bold. Best results among hyperbolic networks with same dimensions are underlined.}
\end{table*}
\egroup

\section{Experiments}
\label{sec:experiments}

To verify our proposed framework, we conduct experiments on both shallow and deep neural networks. For shallow neural networks, we present results on knowledge graph completion. For deep neural networks, we propose a Lorentz Transformer and present results on machine translation. Dependency tree probing is also done on both Lorentz and Euclidean Transformers to compare their capabilities of representing structured information. Due to space limitations, we report the results of network embedding and fine-grained entity typing experiments in the appendix~\ref{sec:appendix-experiments}. For knowledge graph completion and network embedding, we use our fully hyperbolic linear layer, and for other tasks, we use the general formula given in~\cref{sec:components-linear}, which is a relaxation of our fully hyperbolic linear layer.

In the following sections, we denote the models built with our proposed framework as \textsc{\textbf{HyboNet}}. We demonstrate that HyboNet not only outperforms Euclidean and \poincare models on the majority of tasks, but also converges better than its \poincare counterpart. All models in \cref{sec:exp-shallow-networks} are trained with $1$ NVIDIA 2080Ti, models in \cref{sec:exp-deep-networks} are trained with $1$ NVIDIA 40GB A100 GPU. We optimize our model with Riemannian Adam~\cite{geoopt2020kochurov}. For pre-processing and hyper-parameters of each experiment, please refer to Appendix~\ref{sec:appendix-preprocess}.

\subsection{Experiments on Shallow Networks}
\label{sec:exp-shallow-networks}
In this part, we leverage our Lorentz embedding and linear layers to build shallow neural networks. We show that HyboNet outperforms previous knowledge graph completion models on several popular benchmarks.

\subsubsection{Knowledge Graph Completion Models}

A knowledge graph contains a collection of factual triplets, each triplet ($h,r,t$) illustrates the existence of a relation $r$ between the head entity $h$ and the tail entity $t$. Since knowledge graphs are generally incomplete, predicting missing triplets becomes a fundamental research problem. Concretely, the task aims to solve the problem ($h,r,?$) and ($?,r,t$). 

We use two popular knowledge graph completion benchmarks, FB15k-237\cite{toutanova2015observed} and WN18RR\cite{dettmers2018convolutional} in our experiments. We report two evaluation metrics: \textbf{MRR} (Mean reciprocal rank), the average of the inverse of the true entity ranking in the prediction; \textbf{H@$\mathbf{K}$}, the percentage of the correct entities appearing within the top $K$ positions of the predicted ranking.

\begin{figure}[t!]
\centering
\subfloat[WN18RR]{
  \includegraphics[width=0.48\linewidth]{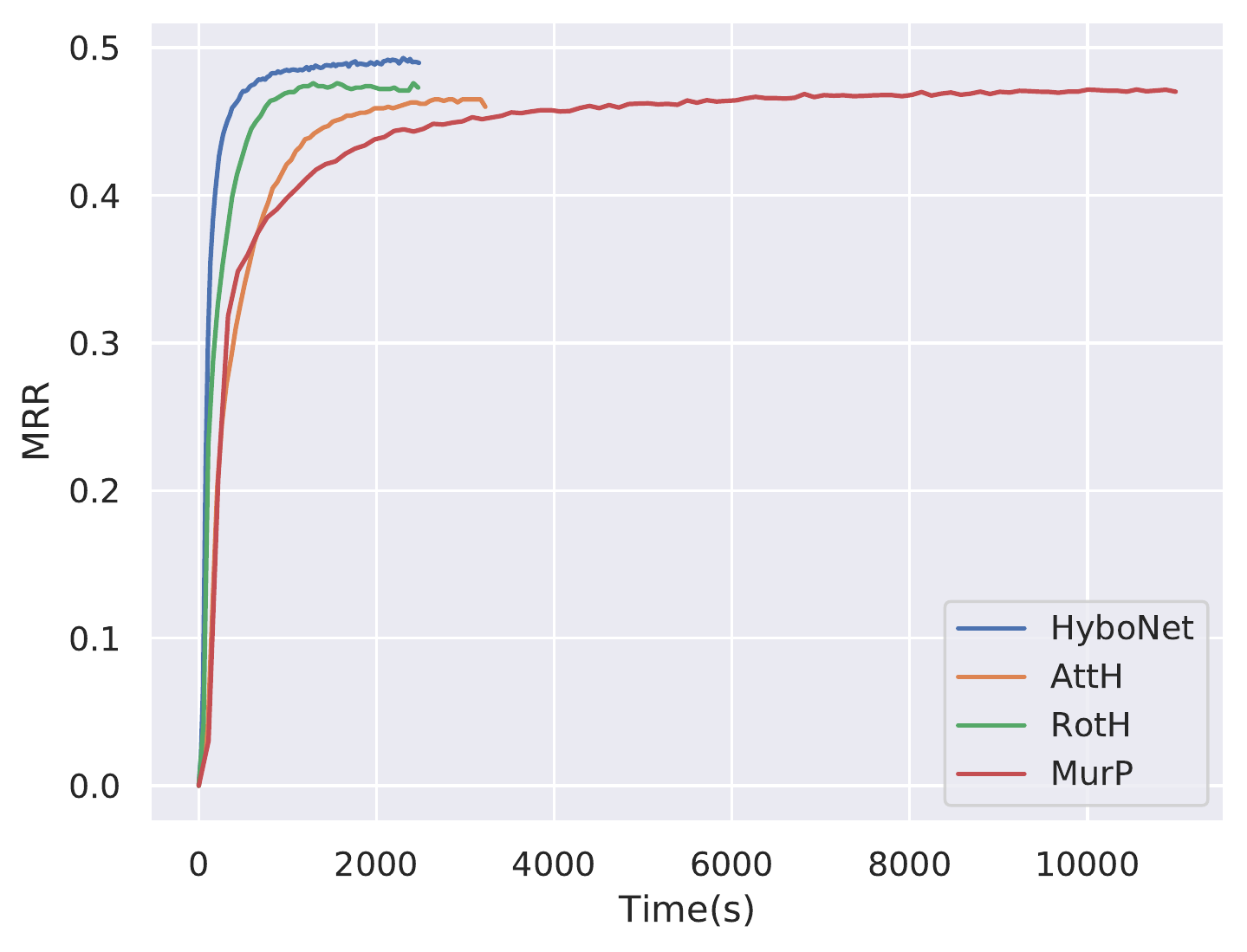}
   \label{fig:mrr-wn18rr}
}
\subfloat[FB15k-237]{
  \includegraphics[width=0.48\linewidth]{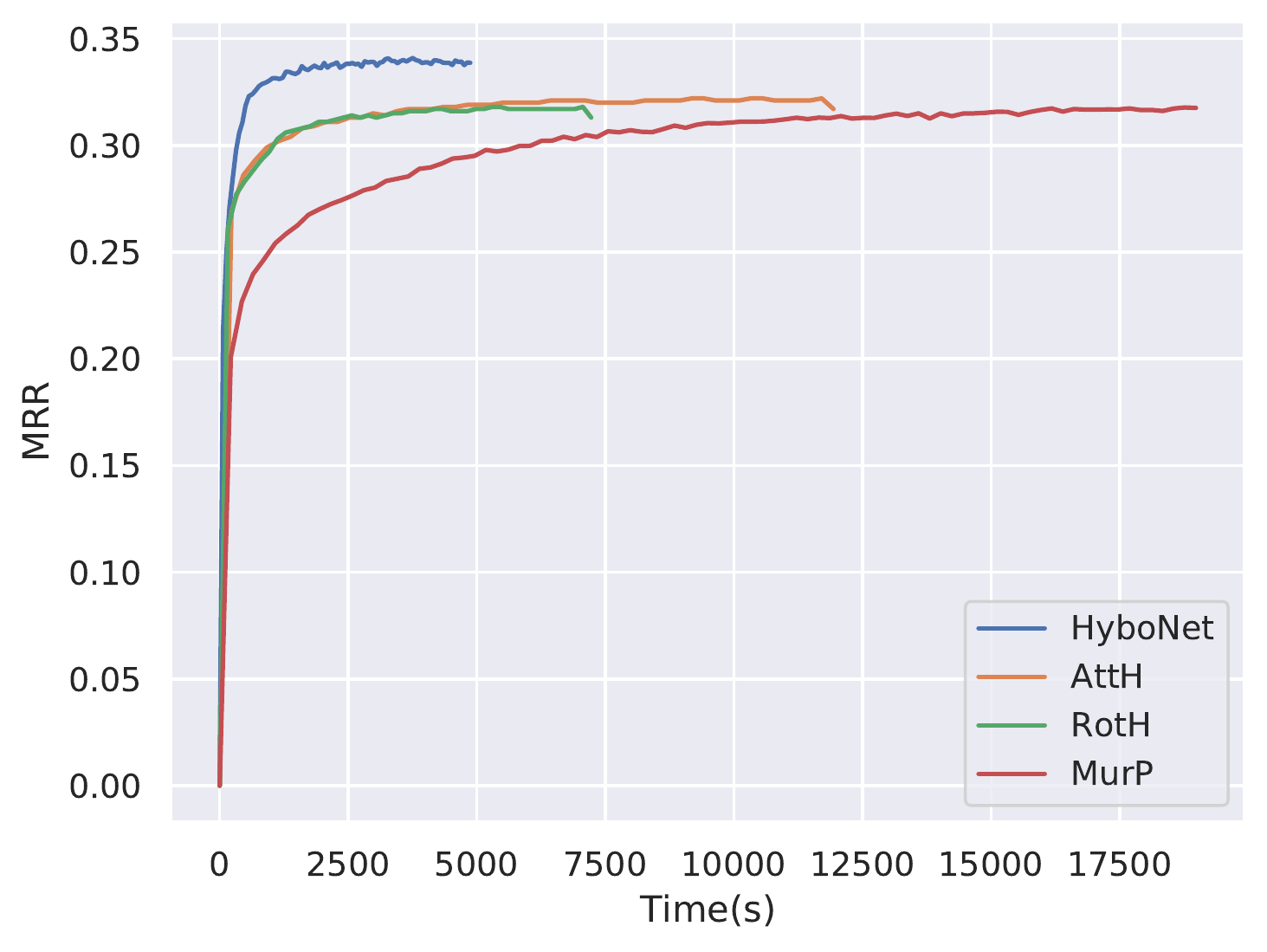}
   \label{fig:mrr-fb15k}
}
\caption{Validation curves of knowledge graph models.}
\end{figure}


\textbf{Setup}\quad Similar to \citet{balazevic2019multi}, we design a score function for each triplet as
\begin{equation*}
\begin{aligned}
\small
s(h,r,t)
&=-d_\mathcal{L}^2(f_r(\mathbf{e}_h), \mathbf{e}_t)+b_h+b_t+\delta,
\end{aligned}
\end{equation*}
where $\mathbf{e}_h,\mathbf{e}_t\in\lorentz$ are the Lorentz embeddings of the head entity $h$ and the tail entity $t$, $f_r(\cdot)$ is a Lorentz linear transformation of the relation $r$ and $\delta$ is a margin hyper-parameter. For each triplet, we randomly corrupt its head or tail entity with $k$ entities and calculate the probabilities for triplets as $p=\sigma(s(h,r,t))$, where $\sigma$ is the sigmoid function. Finally, we minimize the binary cross entropy loss
\begin{equation*}
\small
\mathcal{L} = -\frac{1}{N}\sum_{i=1}^N\left(\log p^{(i)}+\sum_{j=1}^k\log(1-\tilde{p}^{(i,j)})\right),
\end{equation*}
where $p^{(i)}$ and $\tilde{p}^{(i,j)}$ are the probabilities for correct and corrupted triplets respectively, $N$ is the triplet number. We select the model with the best MRR on validation set and report its performance on test set. 

\textbf{Results}\quad \cref{tab:exp-graph-link-prediction} shows the results on both datasets. As expected, low-dimensional hyperbolic networks have already achieved comparable or even better results when compared to high-dimensional Euclidean baselines. When the dimensionality is raised to a maximum of 500, \textsc{HyboNet} outperforms all other baselines on MRR, H@3, and H@1 by a significant margin. And as shown in \cref{fig:mrr-wn18rr,fig:mrr-fb15k}, \textsc{HyboNet} converges better than other hyperbolic networks on both datasets and has a higher ceiling, demonstrating the superiority of our Lorentz linear layer over conventional linear layer formalized in tangent space.


\subsection{Experiments on Deep Networks}
\label{sec:exp-deep-networks}

In this part, we build a Transformer~\cite{vaswani2017attention} with our Lorentz components introduced in \cref{sec:components}. We omit layer normalization for the difficulty of defining hyperbolic mean and variance, but it is still kept in our Euclidean Transformer baseline. In fact, $\lambda$ in Eq.\eqref{eq:hlinear} controls the scaling range, which normalize the representations to some extent. 

\subsubsection{Machine Translation}
\label{sec:exp-mt}

We conduct the experiment on two widely-used machine translation benchmarks: IWSLT'14 English-German and WMT'14 English-German. 

\paragraph{Setup}
We use OpenNMT~\cite{klein-etal-2017-opennmt} to build Euclidean Transformer and our Lorentz one. Following previous hyperbolic work~\cite{shimizu2021hyperbolic}, we conduct experiments in low-dimensional settings. To show that our framework can be applied to high-dimensional settings, we additionally train a Lorentz Transformer of the same size as Transformer base, and compare their performance on WMT'14. We select the model with the lowest perplexity on the validation set, and report its BLEU scores on the test set.

\paragraph{Results}
The BLEU scores on the test set of IWSLT'14 and newstest2013 test set of WMT'14 are shown in \cref{tab:mt-probe}. Both Transformer-based hyperbolic models, \textsc{HyboNet} and \textsc{HAtt}~\cite{gulcehre2018hyperbolic}, outperform the Euclidean Transformer. However, in \textsc{HAtt}, only the calculation of attention weights and the aggregation are performed in hyperbolic space, leaving the remaining computational blocks in the Euclidean space. That is, \textsc{HAtt} is a partially hyperbolic Transformer. As a result, the merits of hyperbolic space are not fully exploited. On the contrary, \textsc{HyboNet} performs all its operations in the hyperbolic space, thus better utilizes the hyperbolic space, and achieve significant improvement over both Euclidean and partially hyperbolic Transformer. Apart from the low-dimensional setting that is common in hyperbolic literature, we scale up the model to be the same size as Transformer base (512-dimensional input)~\cite{vaswani2017attention}. We report the results in \cref{tab:mt-high}. \textsc{HyboNet} outperforms \textsc{Transformer} and \textsc{HAtt} with the same model size, and is very close to the much bigger \textsc{Transformer}$_{\text{big}}$. 


\subsubsection{Dependency Tree Probing}
\label{sec:exp-tree-parsing}

\begin{table}[t]
\small
\centering

\begin{tabular}{l c c c c}
\toprule
 & \textbf{IWSLT'14} & \multicolumn{3}{c}{\textbf{WMT'14}} \\
 \cmidrule(lr){2-2} \cmidrule(lr){3-5}
\textbf{Model} & d=64 & d=64 & d=128 & d=256 \\
\midrule
\textsc{ConvSeq2Seq} & 23.6 & 14.9 & 20.0 & 21.8\\
\textsc{Transformer} & 23.0 & 17.0 & 21.7 & 25.1 \\
\midrule
\textsc{HyperNN++} & 22.0 & 17.0 & 19.4 & 21.8\\
\textsc{HAtt} & 23.7 & 18.8 & 22.5 & 25.5 \\
\textsc{HyboNet} & \textbf{25.9} & \textbf{19.7} & \textbf{23.3} & \textbf{26.2}\\
\bottomrule
\end{tabular}
\caption{The BLEU scores on the test set of IWSLT'14 and WMT'14 under the low-dimensional setting.}
\label{tab:mt-probe}
\end{table}

\begin{table}[t]
\small
    \centering
    \begin{tabular}{l c}
        \toprule
        \textbf{Model} & \textbf{WMT'14}\\
        \midrule
        \textsc{Transformer}$_{\text{base}}$~\cite{vaswani2017attention} & 27.3 \\
        \textsc{Transformer}$_{\text{big}}$~\cite{vaswani2017attention} & \underline{28.4} \\
        \midrule
        \textsc{HAtt}$_{\text{base}}$~\cite{gulcehre2018hyperbolic} & 27.5 \\
        \textsc{HyboNet}$_{\text{base}}$ & \textbf{28.2 }\\
        \bottomrule
    \end{tabular}
    \caption{The BLEU scores on the test set of WMT'14 under the high-dimensional setting. The results of \textsc{Transformer} and \textsc{HAtt} are taken from their original paper respectively.}
    \label{tab:mt-high}
\end{table}

In this part, we verify the superiority of \textsc{HyboNet} in capturing latent structured information in unstructured sentences through dependency tree probing. It has been shown that neural networks implicitly embed syntax trees in their intermediate context representations~\cite{hewitt2019structural,raganato2018analysis}. One reason we think \textsc{HyboNet} performs better in machine translation is that it better captures structured information in the sentences. To validate this, we perform a probing on \textsc{Transformer}, \textsc{HAtt} and \textsc{HyboNet} obtained in \cref{sec:exp-mt}. 
We use dependency parsing result of stanza~\cite{qi2020stanza} on IWSLT'14 English corpus as our dataset. The data partition is kept.

\paragraph{Setup} For a fair comparison, we probe all the models in hyperbolic space following \citet{chen2021probing}. Four metrics are reported: \textbf{UUAS} (undirected attachment score), the percentage of undirected edges placed correctly against the gold tree; \textbf{Root\%}, the precision of the model predicting the root of the syntactic tree; \textbf{Dspr.} and \textbf{Nspr.}, the Spearman correlations between true and predicted distances for each word in each sentence, true depth ordering and the predicted ordering, respectively. Please refer to the appendix for details.

\begin{table}[t]
\small
\centering

\begin{tabular}{l c c c c}
\toprule
& \multicolumn{2}{c}{\textbf{Distance}} & \multicolumn{2}{c}{\textbf{Depth}}\\
 \cmidrule(lr){2-3} \cmidrule(lr){4-5} 
\textbf{Model} & UUAS & Dspr.& Root\% & Nspr. \\
\midrule
\textsc{Transformer} & 0.36 & 0.30 & 12 & 0.88\\\midrule
\textsc{HAtt} & 0.50 & 0.64 & 49 & 0.88\\
\textsc{HyboNet} & \textbf{0.59} & \textbf{0.70} & \textbf{64} & \textbf{0.92}\\
\bottomrule
\end{tabular}
\caption{The probing results on dependency tree constructed from the \textsc{Iwslt'14} English corpus.}
\label{tab:probe}
\end{table}

\paragraph{Results} The probing results are shown in \cref{tab:mt-probe}. \textsc{HyboNet} outperforms other baselines by a large margin. Obviously, syntax trees can be better reconstructed from the intermediate representation of \textsc{HyboNet}'s encoder, which shows that \textsc{HyboNet} better captures syntax structure. The result of \textsc{HAtt} is also worth noting. Because \textsc{HAtt} is a partially hyperbolic Transformer, intuitively, its ability to capture the structured information should be better than Euclidean Transformer, but worse than \textsc{HyboNet}. Our result confirms this suspicion indeed. The probing on \textsc{HAtt} indicates that as the model becomes more hyperbolic, the ability to learn structured information becomes stronger. 

\section{Related Work}

Hyperbolic geometry has been widely investigated in representation learning in recent years, due to its great expression capacity in modeling complex data with non-Euclidean properties. Previous works have shown that when handling data with hierarchy, hyperbolic embedding has better representation capacity and generalization ability~\cite{cvetkovski2016multidimensional,verbeek2014metric,walter2004h,kleinberg2007geographic,krioukov2009curvature,cvetkovski2009hyperbolic,shavitt2008hyperbolic,sarkar2011low}. Moreover, \citet{ganea2018hyperbolic} and \citet{nickel2018learning} introduce the basic operations of neural networks in the \poincare ball and the Lorentz model respectively. After that, researchers further introduce various types of neural models in hyperbolic space including hyperbolic attention networks~\citep{gulcehre2018hyperbolic}, hyperbolic graph neural networks~\citep{liu2019hyperbolic,chami2019hyperbolic}, hyperbolic prototypical networks~\citep{mettes2019hyperspherical} and hyperbolic capsule networks~\citep{chen-etal-2020-hyperbolic}. Recently, with the rapid development of hyperbolic neural networks, people attempt to utilize them in various downstream tasks such as word embeddings~\citep{tifrea2018poincare}, knowledge graph embeddings~\citep{chami-etal-2020-low}, entity typing~\cite{lopez-etal-2019-fine}, text classification~\citep{zhu2020hypertext}, question answering~\citep{tay2018hyperbolic} and machine translation~\citep{gulcehre2018hyperbolic,shimizu2021hyperbolic}, to handle their non-Euclidean properties, and have achieved significant and consistent improvement. 

Our work not only focus on the improvement in the downstream tasks that hyperbolic space offers, but also show that hyperbolic linear transformation used in previous work is just a relaxation of Lorentz rotation, giving a different theoretical interpretation for the hyperbolic linear transformation.

\section{Conclusion and Future Work}

In this work, we propose a novel fully hyperbolic framework based on the Lorentz transformations to overcome the problem that hybrid architectures of existing hyperbolic neural networks relied on the tangent space limit network capabilities. The experimental results on several representative NLP tasks show that compared with other hyperbolic networks, \textsc{HyboNet} has faster speed, better convergence, and higher performance. 
In addition, we also observe that some challenging problems require further efforts: 
(1) Though we have verified the effectiveness of fully hyperbolic models in NLP, exploring its applications in computer vision is still a valuable direction. (2) Though \textsc{HyboNet} has better performance on many tasks, it is slower than Euclidean networks. Also, because of the floating-point error, \textsc{HyboNet} cannot be sped up with half precision training. We hope more efforts can be devoted into this promising field.

\section*{Acknowledgement}
This work is supported by the National Key R\&D Program of China (No. 2020AAA0106502), Institute for Guo Qiang at Tsinghua University, Beijing Academy of Artificial Intelligence (BAAI), and International Innovation Center of Tsinghua University, Shanghai, China.


\bibliography{anthology,custom}

\begin{thebibliography}{61}
\expandafter\ifx\csname natexlab\endcsname\relax\def\natexlab#1{#1}\fi

\bibitem[{Adcock et~al.(2013)Adcock, Sullivan, and Mahoney}]{adcock2013tree}
Aaron~B Adcock, Blair~D Sullivan, and Michael~W Mahoney. 2013.
\newblock Tree-like structure in large social and information networks.
\newblock In \emph{Proceedings of ICDM}, pages 1--10. IEEE Computer Society.

\bibitem[{Balazevic et~al.(2019{\natexlab{a}})Balazevic, Allen, and
  Hospedales}]{balazevic2019multi}
Ivana Balazevic, Carl Allen, and Timothy Hospedales. 2019{\natexlab{a}}.
\newblock Multi-relational poincar{\'e} graph embeddings.
\newblock In \emph{Proceedings of NeurIPS}, pages 4463--4473.

\bibitem[{Balazevic et~al.(2019{\natexlab{b}})Balazevic, Allen, and
  Hospedales}]{balazevic2019tucker}
Ivana Balazevic, Carl Allen, and Timothy Hospedales. 2019{\natexlab{b}}.
\newblock Tucker: Tensor factorization for knowledge graph completion.
\newblock In \emph{Proceedings of EMNLP-IJCNLP}, pages 5188--5197.

\bibitem[{Bordes et~al.(2013)Bordes, Usunier, Garcia-Dur{\'a}n, Weston, and
  Yakhnenko}]{bordes2013translating}
Antoine Bordes, Nicolas Usunier, Alberto Garcia-Dur{\'a}n, Jason Weston, and
  Oksana Yakhnenko. 2013.
\newblock Translating embeddings for modeling multi-relational data.
\newblock In \emph{Proceedings of ICONIP}, pages 2787--2795.

\bibitem[{Bose et~al.(2020)Bose, Smofsky, Liao, Panangaden, and
  Hamilton}]{bose2020latent}
Joey Bose, Ariella Smofsky, Renjie Liao, Prakash Panangaden, and Will Hamilton.
  2020.
\newblock Latent variable modelling with hyperbolic normalizing flows.
\newblock In \emph{Proceedings of ICML}, pages 1045--1055. PMLR.

\bibitem[{Chami et~al.(2020{\natexlab{a}})Chami, Wolf, Juan, Sala, Ravi, and
  R{\'e}}]{chami2020low}
Ines Chami, Adva Wolf, Da-Cheng Juan, Frederic Sala, Sujith Ravi, and
  Christopher R{\'e}. 2020{\natexlab{a}}.
\newblock Low-dimensional hyperbolic knowledge graph embeddings.
\newblock In \emph{Proceedings of ACL}, pages 6901--6914.

\bibitem[{Chami et~al.(2020{\natexlab{b}})Chami, Wolf, Juan, Sala, Ravi, and
  R{\'e}}]{chami-etal-2020-low}
Ines Chami, Adva Wolf, Da-Cheng Juan, Frederic Sala, Sujith Ravi, and
  Christopher R{\'e}. 2020{\natexlab{b}}.
\newblock Low-dimensional hyperbolic knowledge graph embeddings.
\newblock In \emph{Proceedings of ACL}, pages 6901--6914.

\bibitem[{Chami et~al.(2019)Chami, Ying, R{\'e}, and
  Leskovec}]{chami2019hyperbolic}
Ines Chami, Zhitao Ying, Christopher R{\'e}, and Jure Leskovec. 2019.
\newblock Hyperbolic graph convolutional neural networks.
\newblock In \emph{Proceedings of NeurIps}, pages 4869--4880.

\bibitem[{Chen et~al.(2021)Chen, Fu, Xu, Xie, Tan, Chen, and
  Jing}]{chen2021probing}
Boli Chen, Yao Fu, Guangwei Xu, Pengjun Xie, Chuanqi Tan, Mosha Chen, and
  Liping Jing. 2021.
\newblock Probing {\{}bert{\}} in hyperbolic spaces.
\newblock In \emph{Proceedings of ICLR}.

\bibitem[{Chen et~al.(2020)Chen, Huang, Xiao, and
  Jing}]{chen-etal-2020-hyperbolic}
Boli Chen, Xin Huang, Lin Xiao, and Liping Jing. 2020.
\newblock Hyperbolic capsule networks for multi-label classification.
\newblock In \emph{Proceedings of ACL}, pages 3115--3124.

\bibitem[{Choi et~al.(2018)Choi, Levy, Choi, and Zettlemoyer}]{choi2018ultra}
Eunsol Choi, Omer Levy, Yejin Choi, and Luke Zettlemoyer. 2018.
\newblock Ultra-fine entity typing.
\newblock In \emph{Proceedings of ACL}, pages 87--96.

\bibitem[{Cvetkovski and Crovella(2009)}]{cvetkovski2009hyperbolic}
Andrej Cvetkovski and Mark Crovella. 2009.
\newblock Hyperbolic embedding and routing for dynamic graphs.
\newblock In \emph{IEEE INFOCOM 2009}, pages 1647--1655. IEEE.

\bibitem[{Cvetkovski and Crovella(2016)}]{cvetkovski2016multidimensional}
Andrej Cvetkovski and Mark Crovella. 2016.
\newblock Multidimensional scaling in the poincare disk.
\newblock \emph{Applied mathematics \& information sciences}.

\bibitem[{Dettmers et~al.(2018)Dettmers, Minervini, Stenetorp, and
  Riedel}]{dettmers2018convolutional}
Tim Dettmers, Pasquale Minervini, Pontus Stenetorp, and Sebastian Riedel. 2018.
\newblock Convolutional 2d knowledge graph embeddings.
\newblock In \emph{Proceedings of AAAI}.

\bibitem[{Ganea et~al.(2018)Ganea, B{\'e}cigneul, and
  Hofmann}]{ganea2018hyperbolic}
Octavian Ganea, Gary B{\'e}cigneul, and Thomas Hofmann. 2018.
\newblock Hyperbolic neural networks.
\newblock In \emph{Proceedings of NeurIPS}, pages 5345--5355.

\bibitem[{Gu et~al.(2019)Gu, Sala, Gunel, and Ré}]{gu2018learning}
Albert Gu, Frederic Sala, Beliz Gunel, and Christopher Ré. 2019.
\newblock \href {https://openreview.net/forum?id=HJxeWnCcF7} {Learning
  mixed-curvature representations in product spaces}.
\newblock In \emph{Proceedings of ICLR 2019}.

\bibitem[{Gulcehre et~al.(2018)Gulcehre, Denil, Malinowski, Razavi, Pascanu,
  Hermann, Battaglia, Bapst, Raposo, Santoro et~al.}]{gulcehre2018hyperbolic}
Caglar Gulcehre, Misha Denil, Mateusz Malinowski, Ali Razavi, Razvan Pascanu,
  Karl~Moritz Hermann, Peter Battaglia, Victor Bapst, David Raposo, Adam
  Santoro, et~al. 2018.
\newblock Hyperbolic attention networks.
\newblock In \emph{Proceedings of ICLR}.

\bibitem[{Hamilton et~al.(2017)Hamilton, Ying, and
  Leskovec}]{hamilton2017inductive}
William~L Hamilton, Rex Ying, and Jure Leskovec. 2017.
\newblock Inductive representation learning on large graphs.
\newblock In \emph{Proceedings of NeurIPS}, pages 1025--1035.

\bibitem[{Hewitt and Manning(2019)}]{hewitt2019structural}
John Hewitt and Christopher~D Manning. 2019.
\newblock A structural probe for finding syntax in word representations.
\newblock In \emph{Proceedings of NAACL}, pages 4129--4138.

\bibitem[{Jonckheere et~al.(2008)Jonckheere, Lohsoonthorn, and
  Bonahon}]{jonckheere2008scaled}
Edmond Jonckheere, Poonsuk Lohsoonthorn, and Francis Bonahon. 2008.
\newblock Scaled gromov hyperbolic graphs.
\newblock \emph{Journal of Graph Theory}, 57(2):157--180.

\bibitem[{Karcher(1977)}]{https://doi.org/10.1002/cpa.3160300502}
H.~Karcher. 1977.
\newblock \href {https://doi.org/https://doi.org/10.1002/cpa.3160300502}
  {Riemannian center of mass and mollifier smoothing}.
\newblock \emph{Communications on Pure and Applied Mathematics},
  30(5):509--541.

\bibitem[{Kipf and Welling(2017)}]{kipf2017semi}
Thomas~N. Kipf and Max Welling. 2017.
\newblock Semi-supervised classification with graph convolutional networks.
\newblock In \emph{Proceedings of ICLR}.

\bibitem[{Klein et~al.(2017)Klein, Kim, Deng, Senellart, and
  Rush}]{klein-etal-2017-opennmt}
Guillaume Klein, Yoon Kim, Yuntian Deng, Jean Senellart, and Alexander Rush.
  2017.
\newblock {O}pen{NMT}: Open-source toolkit for neural machine translation.
\newblock In \emph{Proceedings of ACL}, pages 67--72.

\bibitem[{Kleinberg(2007)}]{kleinberg2007geographic}
R~Kleinberg. 2007.
\newblock Geographic routing using hyperbolic space.
\newblock In \emph{Proceedings of the IEEE INFOCOM 2007-26th IEEE International
  Conference on Computer Communications}, pages 1902--1909.

\bibitem[{Kochurov et~al.(2020)Kochurov, Karimov, and
  Kozlukov}]{geoopt2020kochurov}
Max Kochurov, Rasul Karimov, and Serge Kozlukov. 2020.
\newblock \href {http://arxiv.org/abs/2005.02819} {Geoopt: Riemannian
  optimization in pytorch}.

\bibitem[{Kolyvakis et~al.(2019)Kolyvakis, Kalousis, and
  Kiritsis}]{kolyvakis2019hyperkg}
Prodromos Kolyvakis, Alexandros Kalousis, and Dimitris Kiritsis. 2019.
\newblock Hyperkg: Hyperbolic knowledge graph embeddings for knowledge base
  completion.
\newblock \emph{arXiv preprint arXiv:1908.04895}.

\bibitem[{Krioukov et~al.(2010)Krioukov, Papadopoulos, Kitsak, Vahdat, and
  Bogun{\'a}}]{krioukov2010hyperbolic}
Dmitri Krioukov, Fragkiskos Papadopoulos, Maksim Kitsak, Amin Vahdat, and
  Mari{\'a}n Bogun{\'a}. 2010.
\newblock Hyperbolic geometry of complex networks.
\newblock \emph{Physical Review E}, 82(3):036106.

\bibitem[{Krioukov et~al.(2009)Krioukov, Papadopoulos, Vahdat, and
  Bogun{\'a}}]{krioukov2009curvature}
Dmitri Krioukov, Fragkiskos Papadopoulos, Amin Vahdat, and Mari{\'a}n
  Bogun{\'a}. 2009.
\newblock Curvature and temperature of complex networks.
\newblock \emph{Physical Review E}, 80(3):035101.

\bibitem[{Law et~al.(2019)Law, Liao, Snell, and Zemel}]{law2019lorentzian}
Marc Law, Renjie Liao, Jake Snell, and Richard Zemel. 2019.
\newblock Lorentzian distance learning for hyperbolic representations.
\newblock In \emph{Proceedings of ICML}, pages 3672--3681.

\bibitem[{Liu et~al.(2019)Liu, Nickel, and Kiela}]{liu2019hyperbolic}
Qi~Liu, Maximilian Nickel, and Douwe Kiela. 2019.
\newblock Hyperbolic graph neural networks.
\newblock In \emph{Proceedings of NeurIPS}, pages 8230--8241.

\bibitem[{L{\'o}pez et~al.(2019)L{\'o}pez, Heinzerling, and
  Strube}]{lopez-etal-2019-fine}
Federico L{\'o}pez, Benjamin Heinzerling, and Michael Strube. 2019.
\newblock Fine-grained entity typing in hyperbolic space.
\newblock In \emph{Proceedings of RepL4NLP}, pages 169--180.

\bibitem[{L{\'o}pez and Strube(2020)}]{lopez2020fully}
Federico L{\'o}pez and Michael Strube. 2020.
\newblock A fully hyperbolic neural model for hierarchical multi-class
  classification.
\newblock In \emph{Proceedings of EMNLP Findings}, pages 460--475.

\bibitem[{Lou et~al.(2020)Lou, Katsman, Jiang, Belongie, Lim, and
  De~Sa}]{lou2020differentiating}
Aaron Lou, Isay Katsman, Qingxuan Jiang, Serge Belongie, Ser-Nam Lim, and
  Christopher De~Sa. 2020.
\newblock Differentiating through the fr{\'e}chet mean.
\newblock In \emph{International Conference on Machine Learning}, pages
  6393--6403. PMLR.

\bibitem[{Mathieu et~al.(2019)Mathieu, Le~Lan, Maddison, Tomioka, and
  Teh}]{mathieu2019continuous}
Emile Mathieu, Charline Le~Lan, Chris~J Maddison, Ryota Tomioka, and Yee~Whye
  Teh. 2019.
\newblock Continuous hierarchical representations with poincar{\'e} variational
  auto-encoders.
\newblock In \emph{Proceedings of NeurIPS}, pages 12565--12576.

\bibitem[{Mettes et~al.(2019)Mettes, van~der Pol, and
  Snoek}]{mettes2019hyperspherical}
Pascal Mettes, Elise van~der Pol, and Cees Snoek. 2019.
\newblock Hyperspherical prototype networks.
\newblock In \emph{Proceedings of NeurIPS}, pages 1487--1497.

\bibitem[{Moretti(2002)}]{moretti2002interplay}
Valter Moretti. 2002.
\newblock The interplay of the polar decomposition theorem and the lorentz
  group.
\newblock \emph{arXiv preprint math-ph/0211047}.

\bibitem[{Narayan and Saniee(2011)}]{narayan2011large}
Onuttom Narayan and Iraj Saniee. 2011.
\newblock Large-scale curvature of networks.
\newblock \emph{Physical Review E}, 84(6):066108.

\bibitem[{Nickel and Kiela(2017)}]{nickel2017poincare}
Maximillian Nickel and Douwe Kiela. 2017.
\newblock Poincar{\'e} embeddings for learning hierarchical representations.
\newblock In \emph{Proceedings of NeurIPS}, pages 6338--6347.

\bibitem[{Nickel and Kiela(2018)}]{nickel2018learning}
Maximillian Nickel and Douwe Kiela. 2018.
\newblock Learning continuous hierarchies in the lorentz model of hyperbolic
  geometry.
\newblock In \emph{Proceedings of ICML}, pages 3779--3788.

\bibitem[{Qi et~al.(2020)Qi, Zhang, Zhang, Bolton, and Manning}]{qi2020stanza}
Peng Qi, Yuhao Zhang, Yuhui Zhang, Jason Bolton, and Christopher~D. Manning.
  2020.
\newblock Stanza: A {Python} natural language processing toolkit for many human
  languages.
\newblock In \emph{Proceedings of ACL}.

\bibitem[{Raganato et~al.(2018)Raganato, Tiedemann
  et~al.}]{raganato2018analysis}
Alessandro Raganato, J{\"o}rg Tiedemann, et~al. 2018.
\newblock An analysis of encoder representations in transformer-based machine
  translation.
\newblock In \emph{Proceedings of EMNLP Workshop}. The Association for
  Computational Linguistics.

\bibitem[{Ramsay and Richtmyer(1995)}]{ramsay1995introduction}
Arlan Ramsay and Robert~D Richtmyer. 1995.
\newblock \emph{Introduction to hyperbolic geometry}.
\newblock Springer Science \& Business Media.

\bibitem[{Sarkar(2011)}]{sarkar2011low}
Rik Sarkar. 2011.
\newblock Low distortion delaunay embedding of trees in hyperbolic plane.
\newblock In \emph{International Symposium on Graph Drawing}, pages 355--366.
  Springer.

\bibitem[{Shavitt and Tankel(2008)}]{shavitt2008hyperbolic}
Yuval Shavitt and Tomer Tankel. 2008.
\newblock Hyperbolic embedding of internet graph for distance estimation and
  overlay construction.
\newblock \emph{IEEE/ACM Transactions on Networking}, 16(1):25--36.

\bibitem[{Shimizu et~al.(2021)Shimizu, Mukuta, and
  Harada}]{shimizu2021hyperbolic}
Ryohei Shimizu, YUSUKE Mukuta, and Tatsuya Harada. 2021.
\newblock Hyperbolic neural networks++.
\newblock In \emph{Proceedings of ICLR}.

\bibitem[{Sun et~al.(2019)Sun, Deng, Nie, and Tang}]{sun2019rotate}
Zhiqing Sun, Zhi-Hong Deng, Jian-Yun Nie, and Jian Tang. 2019.
\newblock Rotate: Knowledge graph embedding by relational rotation in complex
  space.
\newblock In \emph{Proceedings of ICLR}.

\bibitem[{Tay et~al.(2018)Tay, Tuan, and Hui}]{tay2018hyperbolic}
Yi~Tay, Luu~Anh Tuan, and Siu~Cheung Hui. 2018.
\newblock Hyperbolic representation learning for fast and efficient neural
  question answering.
\newblock In \emph{Proceedings of WSDM}, pages 583--591.

\bibitem[{Tifrea et~al.(2018)Tifrea, Becigneul, and Ganea}]{tifrea2018poincare}
Alexandru Tifrea, Gary Becigneul, and Octavian-Eugen Ganea. 2018.
\newblock Poincare glove: Hyperbolic word embeddings.
\newblock In \emph{Proceedings of ICLR}.

\bibitem[{Toutanova and Chen(2015)}]{toutanova2015observed}
Kristina Toutanova and Danqi Chen. 2015.
\newblock Observed versus latent features for knowledge base and text
  inference.
\newblock In \emph{Proceedings of CVSC Workshop}, pages 57--66.

\bibitem[{Trouillon et~al.(2017)Trouillon, Dance, Gaussier, Welbl, Riedel, and
  Bouchard}]{trouillon2017knowledge}
Th{\'e}o Trouillon, Christopher~R Dance, {\'E}ric Gaussier, Johannes Welbl,
  Sebastian Riedel, and Guillaume Bouchard. 2017.
\newblock Knowledge graph completion via complex tensor factorization.
\newblock \emph{The Journal of Machine Learning Research}, 18(1):4735--4772.

\bibitem[{Ungar(2005)}]{ungar2005analytic}
Abraham~A Ungar. 2005.
\newblock \emph{Analytic hyperbolic geometry: Mathematical foundations and
  applications}.
\newblock World Scientific.

\bibitem[{Vaswani et~al.(2017)Vaswani, Shazeer, Parmar, Uszkoreit, Jones,
  Gomez, Kaiser, and Polosukhin}]{vaswani2017attention}
Ashish Vaswani, Noam Shazeer, Niki Parmar, Jakob Uszkoreit, Llion Jones,
  Aidan~N Gomez, {\L}ukasz Kaiser, and Illia Polosukhin. 2017.
\newblock Attention is all you need.
\newblock In \emph{Proceedings of NeurIPS}, pages 5998--6008.

\bibitem[{Veli{\v{c}}kovi{\'c} et~al.(2018)Veli{\v{c}}kovi{\'c}, Cucurull,
  Casanova, Romero, Lio, and Bengio}]{velickovic2018graph}
Petar Veli{\v{c}}kovi{\'c}, Guillem Cucurull, Arantxa Casanova, Adriana Romero,
  Pietro Lio, and Yoshua Bengio. 2018.
\newblock Graph attention networks.
\newblock In \emph{Proceedings of ICLR}.

\bibitem[{Verbeek and Suri(2014)}]{verbeek2014metric}
Kevin Verbeek and Subhash Suri. 2014.
\newblock Metric embedding, hyperbolic space, and social networks.
\newblock In \emph{Proceedings of the thirtieth annual symposium on
  Computational geometry}, pages 501--510.

\bibitem[{Walter(2004)}]{walter2004h}
J{\"o}rg~A Walter. 2004.
\newblock H-mds: a new approach for interactive visualization with
  multidimensional scaling in the hyperbolic space.
\newblock \emph{Information Systems}, 29(4):273--292.

\bibitem[{Wilson et~al.(2014)Wilson, Hancock, Pekalska, and
  Duin}]{wilson2014spherical}
Richard~C Wilson, Edwin~R Hancock, El{\.z}bieta Pekalska, and Robert~PW Duin.
  2014.
\newblock Spherical and hyperbolic embeddings of data.
\newblock \emph{IEEE transactions on pattern analysis and machine
  intelligence}, 36(11):2255--2269.

\bibitem[{Xiong et~al.(2019)Xiong, Wu, Lei, Yu, Chang, Guo, and
  Wang}]{xiong2019imposing}
Wenhan Xiong, Jiawei Wu, Deren Lei, Mo~Yu, Shiyu Chang, Xiaoxiao Guo, and
  William~Yang Wang. 2019.
\newblock Imposing label-relational inductive bias for extremely fine-grained
  entity typing.
\newblock In \emph{Proceedings of NAACL}, pages 773--784.

\bibitem[{Yang et~al.(2015)Yang, Yih, He, Gao, and Deng}]{yang2014embedding}
Bishan Yang, Wen{-}tau Yih, Xiaodong He, Jianfeng Gao, and Li~Deng. 2015.
\newblock Embedding entities and relations for learning and inference in
  knowledge bases.
\newblock In \emph{Proceedings of ICLR}.

\bibitem[{Zhang et~al.(2021{\natexlab{a}})Zhang, Wang, Shi, Jiang, and
  Ye}]{zhang2021hyperbolic}
Yiding Zhang, Xiao Wang, Chuan Shi, Xunqiang Jiang, and Yanfang~Fanny Ye.
  2021{\natexlab{a}}.
\newblock Hyperbolic graph attention network.
\newblock \emph{IEEE Transactions on Big Data}.

\bibitem[{Zhang et~al.(2021{\natexlab{b}})Zhang, Wang, Shi, Liu, and
  Song}]{zhang2021lorentzian}
Yiding Zhang, Xiao Wang, Chuan Shi, Nian Liu, and Guojie Song.
  2021{\natexlab{b}}.
\newblock Lorentzian graph convolutional networks.
\newblock In \emph{Proceedings of the Web Conference 2021}, pages 1249--1261.

\bibitem[{Zhu et~al.(2020)Zhu, Zhou, Xiao, Jiang, Chen, and
  Liu}]{zhu2020hypertext}
Yudong Zhu, Di~Zhou, Jinghui Xiao, Xin Jiang, Xiao Chen, and Qun Liu. 2020.
\newblock Hypertext: Endowing fasttext with hyperbolic geometry.
\newblock In \emph{Proceedings of EMNLP Findings}, pages 1166--1171.

\end{thebibliography}


\begin{thebibliography}{7}
\providecommand{\natexlab}[1]{#1}
\providecommand{\url}[1]{\texttt{#1}}
\expandafter\ifx\csname urlstyle\endcsname\relax
  \providecommand{\doi}[1]{doi: #1}\else
  \providecommand{\doi}{doi: \begingroup \urlstyle{rm}\Url}\fi

\bibitem[Balazevic et~al.(2019)Balazevic, Allen, and
  Hospedales]{balazevic2019multi}
Balazevic, I., Allen, C., and Hospedales, T.
\newblock Multi-relational poincar{\'e} graph embeddings.
\newblock In \emph{Proceedings of NeurIPS}, pp.\  4463--4473, 2019.

\bibitem[Kingma \& Ba(2015)Kingma and Ba]{DBLP:journals/corr/KingmaB14}
Kingma, D.~P. and Ba, J.
\newblock Adam: A method for stochastic optimization.
\newblock In \emph{Proceedings of ICLR}, 2015.
\newblock URL \url{http://arxiv.org/abs/1412.6980}.

\bibitem[Klein et~al.(2017)Klein, Kim, Deng, Senellart, and
  Rush]{klein-etal-2017-opennmt}
Klein, G., Kim, Y., Deng, Y., Senellart, J., and Rush, A.
\newblock {O}pen{NMT}: Open-source toolkit for neural machine translation.
\newblock In \emph{Proceedings of ACL}, pp.\  67--72, 2017.

\bibitem[Kochurov et~al.(2020)Kochurov, Karimov, and
  Kozlukov]{geoopt2020kochurov}
Kochurov, M., Karimov, R., and Kozlukov, S.
\newblock Geoopt: Riemannian optimization in pytorch, 2020.

\bibitem[L{\'o}pez \& Strube(2020)L{\'o}pez and Strube]{lopez2020fully}
L{\'o}pez, F. and Strube, M.
\newblock A fully hyperbolic neural model for hierarchical multi-class
  classification.
\newblock In \emph{Proceedings of EMNLP Findings}, pp.\  460--475, 2020.

\bibitem[Tifrea et~al.(2018)Tifrea, Becigneul, and Ganea]{tifrea2018poincare}
Tifrea, A., Becigneul, G., and Ganea, O.-E.
\newblock Poincare glove: Hyperbolic word embeddings.
\newblock In \emph{Proceedings of ICLR}, 2018.

\bibitem[Vaswani et~al.(2017)Vaswani, Shazeer, Parmar, Uszkoreit, Jones, Gomez,
  Kaiser, and Polosukhin]{vaswani2017attention}
Vaswani, A., Shazeer, N., Parmar, N., Uszkoreit, J., Jones, L., Gomez, A.~N.,
  Kaiser, {\L}., and Polosukhin, I.
\newblock Attention is all you need.
\newblock In \emph{Proceedings of NeurIPS}, pp.\  5998--6008, 2017.

\end{thebibliography}
\bibliographystyle{acl_natbib}

\appendix
\label{sec:appendix}

\section{Other Experiments}
\label{sec:appendix-experiments}

\subsection{Graph Neural Networks}
\label{sec:gnn}

Previous works have shown that when equipped with hyperbolic geometry, GNNs demonstrate impressive improvements compared with its Euclidean counterparts~\cite{chami2019hyperbolic,liu2019hyperbolic}. In this part, we extend GCNs with our proposed hyperbolic framework. Following~\citet{chami2019hyperbolic}, we evaluate our \textsc{HyboNet} for link prediction and node classification on four network embedding datasets, and observe better or comparable results as compared to previous methods.


\paragraph{Setup}
The architecture of GCNs can be summarized into three parts: feature transformation, neighborhood aggregation and non-linear activation. We use a Lorentz linear layer for the feature transformation, and use the centroid of neighboring node features as the aggregation result. The non-linear activation is integrated into Lorentz linear layer as elaborated in \cref{sec:components-linear}. The overall operations of the $l$-th network layer can be formulated into the following manner:
\begin{equation*}
\small
    \mathbf{x}_i^l=\texttt{Att}(\texttt{HL}(\mathbf{x}_i^{l-1}), \{\texttt{HL}(\mathbf{x}_{j\in\mathcal{N}(i)}^{l-1})\}, \{\texttt{HL}(\mathbf{x}_{j\in\mathcal{N}(i)}^{l-1})\}),
\end{equation*}
where $\mathbf{x}_i^l$ refers to the representation of the $i$-th node at the layer $l$, $\mathcal{N}(i)$ denotes the neighboring nodes of the $i$-th node. With the node representation, we can easily conduct link prediction and node classification. For link prediction, we calculate the probability of edges using Fermi-Dirac decoder~\cite{krioukov2010hyperbolic,nickel2017poincare}:
\begin{equation}
\small
    p((i,j)\in\mathcal{E}\mid\mathbf{x}_i, \mathbf{x}_j)=\left(\exp((d^2_\mathcal{L}(\mathbf{x}_i, \mathbf{x}_j)-r)/t)+1\right)^{-1},
\end{equation}
where $r$ and $t$ are hyper-parameters. We minimize the binary cross entropy loss. For node classification, we calculate the squared Lorentzian distance between node representation and class representations, and minimize the cross entropy loss.


\bgroup
\setlength{\tabcolsep}{4.5pt}
\def\arraystretch{0.9}
\begin{table*}[t]
\small
\begin{center}

  \begin{tabular}{l*{8}{c}}
    \toprule
    & \multicolumn{2}{c}{\textbf{Disease}($\delta=0$)} & \multicolumn{2}{c}{\textbf{Airport}($\delta=1$)} & \multicolumn{2}{c}{\textbf{PubMed}($\delta=3.5$)} & \multicolumn{2}{c}{\textbf{Cora}($\delta=11$)} \\
    \cmidrule(lr){2-3}\cmidrule(lr){4-5}\cmidrule(lr){6-7}\cmidrule(lr){8-9}
    \textbf{Task} & \multicolumn{1}{c}{LP} & \multicolumn{1}{c}{NC} & \multicolumn{1}{c}{LP} & \multicolumn{1}{c}{NC} & \multicolumn{1}{c}{LP} & \multicolumn{1}{c}{NC} & \multicolumn{1}{c}{LP} & \multicolumn{1}{c}{NC}\\
    \midrule
    \textsc{GCN}~\cite{kipf2017semi} & 64.7$_{\pm0.5}$ & 69.7$_{\pm0.4}$ & 89.3$_{\pm0.4}$ & 81.4$_{\pm0.6}$ & 91.1$_{\pm0.5}$ & 78.1$_{\pm0.2}$ & 90.4$_{\pm0.2}$ & 81.3$_{\pm0.3}$\\
    \textsc{GAT}~\cite{velickovic2018graph} & 69.8$_{\pm0.3}$ & 70.4$_{\pm0.4}$ & 90.5$_{\pm0.3}$ & 81.5$_{\pm0.3}$ & 91.2$_{\pm0.1}$ & \textbf{79.0}$_{\pm0.3}$ & \textbf{93.7}$_{\pm0.1}$ & 83.0$_{\pm0.7}$\\
    \textsc{SAGE}~\cite{hamilton2017inductive} & 65.9$_{\pm0.3}$ & 69.1$_{\pm0.6}$ & 90.4$_{\pm0.5}$ & 82.1$_{\pm0.5}$ & 86.2$_{\pm1.0}$ & 77.4$_{\pm2.2}$ & 85.5$_{\pm0.6}$ & 77.9$_{\pm2.4}$\\
    \textsc{SGC}~\cite{wilson2014spherical} & 65.1$_{\pm0.2}$ & 69.5$_{\pm0.2}$ & 89.8$_{\pm0.3}$ & 80.6$_{\pm0.1}$ & 94.1$_{\pm0.0}$ & 78.9$_{\pm0.0}$ & 91.5$_{\pm0.1}$ & 81.0$_{\pm0.1}$\\\midrule
    \textsc{HGCN}~\cite{chami2019hyperbolic} & 91.2$_{\pm0.6}$ & 82.8$_{\pm0.8}$ & 96.4$_{\pm0.1}$ & 90.6$_{\pm0.2}$ & 96.1$_{\pm0.2}$ & 78.4$_{\pm0.4}$ & 93.1$_{\pm0.4}$ & 81.3$_{\pm0.6}$\\
    \textsc{HAT}~\cite{zhang2021hyperbolic} & 91.8$_{\pm0.5}$ & 83.6$_{\pm0.9}$ & / & / & 96.0$_{\pm0.3}$ & 78.6$_{\pm0.5}$ & 93.0$_{\pm0.3}$ & 83.1$_{\pm0.6}$ \\
    \textsc{LGCN}~\cite{zhang2021lorentzian} & 96.6$_{\pm0.6}$ & 84.4$_{\pm0.8}$ & 96.0$_{\pm0.6}$ & 90.9$_{\pm1.7}$ & \textbf{96.8$_{\pm0.1}$} & 78.6$_{\pm0.7}$ & 93.6$_{\pm0.4}$ & \textbf{83.3}$_{\pm0.7}$\\
    \textsc{HyboNet} & \textbf{96.8}$_{\pm0.4}$ & \textbf{96.0}$_{\pm1.0}$ & \textbf{97.3}$_{\pm0.3}$ & \textbf{90.9}$_{\pm1.4}$ & 95.8$_{\pm0.2}$ & 78.0$_{\pm1.0}$ & 93.6$_{\pm0.3}$ & 80.2$_{\pm1.3}$\\
   \bottomrule
  \end{tabular}
  \label{tab:exp-network-embedding}
\end{center}
\caption{Test ROC AUC results (\%) for Link Prediction (LP) and F1 scores (\%) for Node Classification (NC). \textsc{HGCN} and \textsc{HyboNet} are hyperbolic models. $\delta$ refers to Gromovs $\delta$-hyperbolicity, and is given by~\citet{chami2019hyperbolic}. The lower the $\delta$, the more hyperbolic the graph.}
\end{table*}
\egroup

\paragraph{Results}
Following~\citet{chami2019hyperbolic}, we report ROC AUC results for link prediction and F1 scores for node classification on four different network embedding datasets. The description of the datasets can be found in our appendix. \citet{chami2019hyperbolic} compute Gromovs $\delta$-hyperbolicity\cite{jonckheere2008scaled,adcock2013tree,narayan2011large} for these four datasets. The lower the $\delta$ is, the more hyperbolic the graph is. 

The results are reported in \cref{tab:exp-network-embedding}. \textsc{HyboNet} outperforms other baselines in those highly hyperbolic datasets. For Disease dataset, \textsc{HyboNet} even achieves a 12\% (absolute) improvement on node classification over previous hyperbolic GCNs. 
On the less hyperbolic datasets such as PubMed and Cora, \textsc{HyboNet} still performs well on link prediction, and remains competitive for node classification. Although \textsc{HyboNet} does not significantly better than \textsc{LGCN} on all datasets, we observe that \textsc{HyboNet} is far more stable than \textsc{LGCN}. Out of 128 link prediction experiments in grid search, there are 89 times that \textsc{LGCN} generates NaN and fails to finish training, while \textsc{HyboNet} remains stable and is faster than \textsc{LGCN}.

\subsection{Fine-grained Entity Typing}
\label{sec:exp-typing}

Given a sentence containing a mention of entity $e$, the purpose of entity typing is to predict the type of $e$ from a type inventory. It is a multi-label classification problem since multiple types can be assigned to $e$. For fine-grained entity typing, type labels are divided into finer granularity, making the type inventory contains thousands of types. We conduct the experiment on Open Entity dataset~\cite{choi2018ultra}, which divides types into three levels: coarse, fine, and ultra-fine.

\bgroup
\setlength{\tabcolsep}{1pt}
\begin{table}[t]
\small

  \begin{center}
    \begin{tabular}{lccccr}
      \toprule
      Model & \multicolumn{1}{c}{\textbf{Total}} & \multicolumn{1}{c}{\textbf{C}} & \multicolumn{1}{c}{\textbf{F}} & \multicolumn{1}{c}{\textbf{UF}} & \textbf{\#Para} \\
      \midrule
      \textsc{LabelGCN} & 35.8 & 67.5 & 42.2 & 21.3 & 5.1M\\
      \textsc{MultiTask} & 31.0 & 61.0 & 39.0 & 14.0 & 6.1M\\
      \midrule
       \textsc{HY Base} & 36.3 & \textbf{68.1} & 38.9 & 21.2 & 1.8M \\
      \textsc{HY Large} & 37.4 & 67.6 & 41.4 & 24.7 & 4.6M \\
      \textsc{HY xLarge} & \textbf{38.2} & 67.1 & 40.4 & \underline{\textbf{25.7}} & 9.5M \\
      \midrule
      \textsc{Lorentz (Tangent)} & 37.2 & 68.0 & 40.3 & 22.4 & 2.9M\\
      \textsc{HyboNet} & \textbf{38.2} & \textbf{68.1} & \textbf{43.2} & 23.5 & 2.9M\\
      \bottomrule
    \end{tabular}
    \label{tab:entity-typing}
  \end{center}
  \caption{Macro $F_1$ scores (\%) on the development set of Open Entity dataset for different baselines and models. Best results are underlined, and best results among hyperbolic models are in bold.}
\end{table}
\egroup

\paragraph{Setup} \quad Our entity typing model consists of a mention encoder and a context encoder. To get mention representation, the mention encoder first obtain word representation $\mathbf{s}_i$, then calculate the centroid of $\mathbf{s}_i$ as mention representation $\mathbf{m}$ according to Eq.~\eqref{eq:mass-center} with uniform weight. The context encoder is a Lorentz Transformer encoder that shares the same embedding module with mention encoder. The context representation $\mathbf{c}$ is the distance-based attention~\cite{lopez2020fully} result over the Transformer encoder's output. 
We combine $\mathbf{m}$ and $\mathbf{c}$ in the way of combining multi-headed outputs described in \cref{sec:components-attention} by regarding $\mathbf{m}$ and $\mathbf{c}$ as a two-headed output. We then calculate a probability $p_i = \sigma(-d_\mathcal{L}^2(\mathbf{r}, \mathbf{t}_i) / \alpha_i + \beta_i)$ for every type label,
where $\sigma$ is sigmoid function, $\mathbf{t}_i$ is the Lorentz embedding of the $i$-th type label, and $\alpha_i, \beta_i$ are learnable scale and bias factor respectively. During training, we optimize the multi-task objective~\cite{vaswani2017attention}. For evaluation, a type is predicted if its probability is larger than $0.5$.

\paragraph{Results} \quad Following previous works~\cite{choi2018ultra,lopez2020fully}, we report the macro-averaged F1 scores on the development set of Open Entity dataset in \cref{tab:entity-typing}. HyboNet outperforms LabelGCN~\cite{xiong2019imposing} and MultiTask~\cite{vaswani2017attention} on Total with fewer parameters. Compared with large Euclidean models Denoised and BERT, HyboNet achieves comparable fine and ultra-fine results with significantly fewer parameters. Compared with another hyperbolic model HY~\cite{lopez2020fully}, which is based on the \poincare ball model, HyboNet outperforms HY xLarge model on coarse and fine results. Note that HyboNet has only slightly more parameters than HY base, and fewer than HY Large. 

\section{Data Preprocessing Methods}
\label{sec:appendix-preprocess}
We describe data preprocessing methods for each experiment in this section.
\subsection{Knowledge Graph Completion}

The statistics of WN18RR and FB15k-237 are listed in \cref{tab:statistics-graph}. We keep our data preprocessing method for knowledge graph completion the same as \citet{balazevic2019multi}. Concretely, we augment both WN18RR and FB15k-237 by adding reciprocal relations for every triplet, i.e. for every ($h, r, t$) in the dataset, we add an additional triplet ($t, r^{-1}, h$).

\subsection{Machine Translation}
For WMT'14, we use the preprocessing script provided by OpenNMT\footnote{\url{https://github.com/OpenNMT/OpenNMT-tf/tree/master/scripts/wmt}}. For IWSLT'14, we clean and partition the dataset with script provided by FairSeq\footnote{\url{https://github.com/pytorch/fairseq/tree/master/examples/translation}}. We limit the lengths of both source and target sentences to be 100 and do not share the vocabulary between source and target language. 

\subsection{Network Embedding}
We use four datasets, referred to as Disease, Airport, Pubmed and Cora. The four datasets are preprocessed by \citet{chami2019hyperbolic} and published in their code repository\footnote{\url{https://github.com/HazyResearch/hgcn}}. We refer the readers to \citet{chami2019hyperbolic} for further information about the datasets.

\subsection{Entity Typing}
The dataset consist of $6,000$ crowd sourced samples and $6$M distantly supervised training samples. We keep our data preprocessing method for knowledge graph completion the same as \citet{lopez2020fully}. For the input context, We trimmed the sentence to a maximum of 25 words. During the trimming, one word at a time is removed from one side of the mention, trying to keep the mention in the center of the sentence, and preserve the context information of the mention. For the input mention, we trimmed the mention to a maximum of 5 words. 

\section{Experiment Details}
\label{sec:appendix-detail}

All of our experiments use 32-bit floating point numbers, not 64-bit floating point numbers as in most previous work. We use PyTorch as the neural networks' framework. The negative curvature $K$ of the Lorentz model in our experiments is $-1$.

We take the function $\phi$ in Lorentz linear layer to have the form 
\small
\begin{equation}
    \phi(\mathbf{Wx})=\frac{\sqrt{(\lambda\sigma(\mathbf{v}^T\mathbf{x}+b)+\epsilon)^2+1/K}}{\lVert\mathbf{W}h\big(\texttt{dropout}(\mathbf{x})\big)\rVert}\mathbf{W}h \big(\texttt{dropout}(\mathbf{x}) \big).
    \label{eq:linear}
\end{equation}
\normalsize

To see what it means, we first compute $\mathbf{y}_0=\lambda\sigma(\mathbf{v}^T\mathbf{x}+b)+\epsilon$ as the $0$-th dimension of the output $\mathbf{y}$, where $\sigma$ is the sigmoid function, $\lambda$ controls the $0$-th dimension's range, it can be either learnable or fixed, $b$ is a learnable bias term, and $\epsilon>\sqrt{1/K}$ is a constant preventing the $0$-th dimension be smaller than $\sqrt{1/K}$. According to the definition of Lorentz model,  $\mathbf{y}$ should satisfies $\lVert\mathbf{y}_{1:n}\rVert^2-\mathbf{y_0}^2=1/K$, that is, $\lVert\mathbf{y}_{1:n}\rVert=\sqrt{\mathbf{y_0}^2+1/K}=\sqrt{(\lambda\sigma(\mathbf{v}^T\mathbf{x}+b)+\epsilon)^2+1/K}$. Then \cref{eq:linear} can be seen as first calculate $\tilde{\mathbf{y}}_{1:n}=\mathbf{W}h\big(\texttt{dropout}(\mathbf{x})\big)$, then scale $\tilde{\mathbf{y}}_{1:n}$ to have vector norm $\lVert\mathbf{y}_{1:n}\rVert$ to obtain $\mathbf{y}_{1:n}$. Finally, we concatenate $\mathbf{y_0}$ with $\mathbf{y}_{1:n}$ as output. 

For residual and position embedding addition, we also use Eq.\eqref{eq:linear}.



\subsection{Initialization}

\bgroup
\setlength{\tabcolsep}{4pt}
\begin{table}[t!]
    \centering
\small
  \begin{tabular}{l r r r r r}
    \toprule
    \textbf{Dataset} & \textbf{\#Ent} & \textbf{\#Rel} & \textbf{\#Train} & \textbf{\#Valid} & \textbf{\#Test}\\
    \midrule
    FB15k-237 & 14,541 & 237 & 272,115 & 17,535 & 20,466\\
    WN18RR & 40,943 & 11 & 86,835 & 3,034 & 3,134\\
    \bottomrule
  \end{tabular}
  
  \caption{Statistics of FB15k-237 and WN18RR.}
  \label{tab:statistics-graph}
\end{table}
\egroup

\begin{table}
\small
    \centering
    \setlength{\tabcolsep}{3pt}
    \begin{tabular}{l r}
        \toprule
        \textbf{Embedding}\\
        $\mathbf{x}_i$ & Geoopt default\\
        Parameters in $f$ & \texttt{Uniform}(-0.02, 0.02)\\
        \midrule
        \textbf{Lorentz Linear Layer} & \\
        $\mathbf{W}$ & \texttt{Uniform}(-0.02, 0.02)\\
        $\mathbf{v}$ & \texttt{Uniform}(-0.02, 0.02)\\
        \bottomrule
    \end{tabular}
    \caption{Initialization methods of different parameters.}
    \label{tab:init}
\end{table}
We use different initialization method for different parameters, see Table~\ref{tab:init}. Geoopt\cite{geoopt2020kochurov} initialize the parameter with Gaussian distribution in the tangent space, and map the embedding to hyperbolic space with exponential map. 

\subsection{Knowledge Graph Completion}

\begin{table}[t!]
\small
    \centering
    \centering
    \begin{tabular}{l r r r r}
        \toprule
        & \multicolumn{2}{c}{\textbf{WN18RR}} & \multicolumn{2}{c}{\textbf{FB15k-237}}\\
        Dimension & 32 & 500 & 32 & 500\\
        \midrule
        Batch Size & 1000 & 1000 & 500 & 500\\
        Neg Samples & 50 & 50 & 50 & 50\\
        Margin & 8.0 & 8.0 & 8.0 & 8.0\\
        Epochs & 1000 & 1000 & 500 & 500\\
        Max Norm & 1.5 & 2.5 & 1.5 & 1.5\\
        $\lambda$ & 3.5 & 2.5 & 2.5 & 2.5\\
        Learning Rate & 0.005 & 0.003 & 0.003 & 0.003\\
        Grad Norm & 0.5 & 0.5 & 0.5 & 0.5\\
        Optimizer & rAdam & rAdam & rAdam & rAdam\\
        \bottomrule
    \end{tabular}
    \caption{Hyper-parameters for knowldge graph completion.}
    \label{tab:hyp-graph}
\end{table}

\begin{table*}[t!]
    \centering
    \small
    \begin{tabular}{l r r r r r r r r}
        \toprule
        & \multicolumn{2}{c}{\textbf{Disease}($\delta=0$)} & \multicolumn{2}{c}{\textbf{Airport}($\delta=1$)} & \multicolumn{2}{c}{\textbf{PubMed}($\delta=3.5$)} & \multicolumn{2}{c}{\textbf{Cora}($\delta=11$)} \\
      \cmidrule(lr){2-3}\cmidrule(lr){4-5}\cmidrule(lr){6-7}\cmidrule(lr){8-9}
      \textbf{Task} & \multicolumn{1}{c}{LP} & \multicolumn{1}{c}{NC} & \multicolumn{1}{c}{LP} & \multicolumn{1}{c}{NC} & \multicolumn{1}{c}{LP} & \multicolumn{1}{c}{NC} & \multicolumn{1}{c}{LP} & \multicolumn{1}{c}{NC}\\
      \midrule
      Learning Rate & 0.005 & 0.005 & 0.01 & 0.02 & 0.008 & 0.02 & 0.02 & 0.02\\
      Weight Decay & 0 & 0 & 0.002 & 0.0001 &  0 & 0.001 & 0.001 & 0.01\\
      Dropout & 0.0 & 0.1 & 0.0 & 0.0 & 0.5 & 0.8 & 0.7 & 0.9\\
      Layers & 2 & 4 & 2 & 6 & 2 & 3 & 2 & 3\\
      Max Grad Norm & None & 0.5 & 0.5 & 1 & 0.5 & 0.5 & 0.5 & 1\\
     \bottomrule
        
    \end{tabular}
    \caption{Hyper-parameters for network embeddings.}
    \label{tab:network-embedding}
\end{table*}

We list the hyper-parameters used in the experiment in Table~\ref{tab:hyp-graph}. Note that in this experiment, we restrict the norm of the last $n$ dimension of the embeddings to be no bigger than a certain value, referred to as Max Norm in Table~\ref{tab:hyp-graph}. For each dataset, we explore $\text{BatchSize}\in\{500, 1000\}$, $\text{Margin}\in\{4, 6, 8\}$, $\text{MaxNorm}\in\{1.5, 2.5, 3.5\}$, $\lambda\in\{2.5, 3.5, 5.5\}$, $\text{LearningRate}\in\{3e-3, 5e-3, 7e-3\}$.

\begin{table*}
\small
    \centering
    \begin{tabular}{l r r r r}
        \toprule
        \textbf{Hyper-parameter} & \textbf{IWSLT'14} & \multicolumn{3}{c}{\textbf{WMT'16}}\\
        \midrule
        GPU Numbers & 1 & 4 & 4 & 4\\
        Embedding Dimension & 64 & 64 & 128 & 256\\
        Feed-forward Dimension & 256 & 256 & 512 & 1024\\
        Batch Type & Token & Token & Token & Token\\
        Batch Size Per GPU & 10240 & 10240 & 10240 & 10240\\
        Gradient Accumulation Steps & 1 & 1 & 1 & 1\\
        Training Steps & 40000 & 200000 & 200000 & 200000\\
        Dropout & 0.0 & 0.1 & 0.1 & 0.1\\
        Attention Dropout & 0.1 & 0.0 & 0.0 & 0.0\\
        Max Gradient Norm & 0.5 & 0.5 & 0.5 & 0.5\\
        Warmup Steps & 8000 & 6000 & 6000 & 6000\\
        Decay Method & noam & noam & noam & noam\\
        Label Smoothing & 0.1 & 0.1 & 0.1 & 0.1\\
        Layer Number & 6 & 6 & 6 & 6\\
        Head Number & 4 & 4 & 8 & 8\\
        Learning Rate & 5 & 5 & 5 & 5\\
        Optimizer & rAdam & rAdam & rAdam & rAdam\\
        \bottomrule
    \end{tabular}
    \caption{Hyper-parameters for machine translation.}
    \label{tab:hyp-mt}

\end{table*}

\subsection{Machine Translation}
Our code is based on OpenNMT's Transformer\cite{klein-etal-2017-opennmt}. The hyper-parameters are listed in Table~\ref{tab:hyp-mt}
\subsection{Dependency Tree Probing}
\label{sec:appendix-detail-probing}
The probing for the Euclidean Transformer is done by first applying an Euclidean linear mapping $f_P:\Real^n\rightarrow\Real^{m+1}$ followed by a projection to map Transformer's intermediate context-aware representation $\mathbf{c}_i$ into points $\mathbf{\tilde{h}}_i$ in tangent space of Lorentz model's origin, then using exponential map to map $\mathbf{\tilde{h}}_i$ to hyperbolic space $\mathbf{p}_i$. In the hyperbolic space, we construct the Lorentz syntactic subspace via a Lorentz linear layer $f_Q:\mathbb{L}_K^m\rightarrow\mathbb{L}_K^m$:
\begin{equation*}
\begin{aligned}
  \mathbf{p}_i &= \exp_\mathbf{0}^K(f_P(\mathbf{c}_i)),\\
  \mathbf{q}_i &= f_Q(\mathbf{p}_i).
\end{aligned}
\end{equation*}

We use the squared Lorentzian distance between $\mathbf{q}_i$ and $\mathbf{q}_j$ to recreate tree distances between word pairs $w_i$ and $w_j$, the squared Lorentzian distance between $\mathbf{q}_i$ and the origin $\mathbf{o}$ to recreate the depth of word $w_i$. We minimize the following loss:
\begin{equation*}
\begin{aligned}
  \mathcal{L}_{\text{distance}}&=\frac{1}{l^2}\sum_{i,j\in\{1,\cdots,t\}}|d_T(w_i, w_j)-d_{\mathcal{L}}^2(\mathbf{q}_i, \mathbf{q}_j)|\\
  \mathcal{L}_{\text{depth}}&=\frac{1}{l}\sum_{i\in\{1,\cdots,t\}}|d_D(w_i)-d_{\mathcal{L}}^2(\mathbf{q}_i, \mathbf{o})|,
\end{aligned}
\end{equation*}
where $d_T(w_i,w_j)$ is the edge number of the shortest path from $w_i$ to $w_j$ in the dependency tree, and $l$ is the sentence length. For the probing of Lorentz Transformer, we only substitute $f_P$ with a Lorentz one, and discard the exponential map. We probe every layer for both models, and report the results of the best layer.

We do the probing in the 64 dimensional hyperbolic space. The hyper-parameters and the best layer we choose according to development set are listed in Table~\ref{tab:hyp-probe}. Because no Lorentz embedding is involved, we simply use Adam as the optimizer. For parameter selection, we explore $\text{Learning Rate}\in\{5e-4, 3e-4, 1e-4, 5e-5, 3e-5, 1e-5\}$, $\text{Weight Decay}\in\{0, 1e-6, 1e-5, 1e-4\}$, Batch Size$\in\{16, 32, 64\}$.

\subsection{Network Embedding}
The experiment setting is the same as \citet{chami2019hyperbolic}. We list the hyper-parameters for the four datasets in Table.\ref{tab:network-embedding}

\subsection{Entity Typing}
We initialize the word embeddings by isometrically projecting the pretrained \poincare Glove word embeddings~\cite{tifrea2018poincare} to Lorentz model, and fix them during training. A Lorentz linear layer is applied to transform the word embeddings to a higher dimension. To get mention representation, the mention encoder first obtain word representation $\mathbf{s}_i$ through position encoding module described in section 3.4, then calculate the centroid of $\mathbf{s}_i$ as mention representation $\mathbf{m}_i$ according to Eq. 14 with uniform weight
\begin{equation*}
\begin{aligned}
  \mathbf{s}_i &= \texttt{PE}(\texttt{HL}(\mathbf{w}_i)),\\
  \mathbf{m}_i &= \texttt{Centroid}(1/l, \mathbf{s}_i),
\end{aligned}
\end{equation*}
where $l$ is the length of sentence, $\mathbf{w}_i$ is the pretrained embedding of $i$-th word. The context encoder is a Lorentz Transformer encoder that shares the same embedding module with mention encoder. The context representation $\mathbf{c}$ is the distance-based attention~\cite{lopez2020fully} result over the Transformer encoder's output $\mathbf{h}_i$:
\begin{equation*}
\begin{aligned}
  \mathbf{x}_i = \texttt{PE}(\mathbf{h}_i)&,\quad
  \mathbf{q}_i = \texttt{HL}(\mathbf{x}_i),\quad
  \mathbf{k}_i = \texttt{HL}(\mathbf{x}_i),\\
  \nu_{ij} &= \texttt{Softmax}(-d_{\Lc}^2(\mathbf{q}_i, \mathbf{k}_j) / \sqrt{n}),\\
  \mathbf{c} &= \texttt{Centroid}(\nu_{ij}, \mathbf{h}_i).
\end{aligned}
\end{equation*}

\begin{table}
\small
    \caption{Hyper-parameters for dependency tree probing.}
    \centering
    \begin{tabular}{l r r r}
        \toprule
        Hyper-parameter & Euclidean & HAtt & HyboNet\\
        \midrule
        Learning Rate & 5e-5 & 5e-5 & 5e-5\\
        Weight Decay & 0 & 1e-6 & 0\\
        Best Layer & 0 & 3 & 4\\
        Batch Size & 64 & 32 & 32\\
        Steps & 20000 & 20000 & 20000\\
        Optimizer & Adam & Adam & Adam\\
        \bottomrule
    \end{tabular}
    \label{tab:hyp-probe}

\end{table}

\end{document}


\def\UrlBreaks{\do\/\do-}

\twocolumn[
\icmltitle{Supplementary Material for Fully Hyperbolic Neural Networks}

\vskip 0.3in
]




\appendix

\section{Data Preprocessing Methods}
We describe data preprocessing methods for each experiment in this section.
\subsection{Knowledge Graph Completion}

The statistics of WN18RR and FB15k-237 are listed in \cref{tab:statistics-graph}. We keep our data preprocessing method for knowledge graph completion the same as \citet{balazevic2019multi}. Concretely, we augment both WN18RR and FB15k-237 by adding reciprocal relations for every triplet, i.e. for every ($h, r, t$) in the dataset, we add an additional triplet ($t, r^{-1}, h$).

\subsection{Machine Translation}
For WMT'16, we use the preprocessing script provided by OpenNMT\footnote{\url{https://github.com/OpenNMT/OpenNMT-tf/tree/master/scripts/wmt}}. For IWSLT'14, we clean and partition the dataset with script provided by FairSeq\footnote{\url{https://github.com/pytorch/fairseq/tree/master/examples/translation}}, but we process the corpus with sentencepiece's unigram language model\footnote{\url{https://github.com/google/sentencepiece}} rather than BPE in the original script. We limit the lengths of both source and target sentences to be 100 and share the vocabulary between source and target language. 

\subsection{Entity Typing}
The dataset consist of $6,000$ crowd sourced samples and $6$M distantly supervised training samples. We keep our data preprocessing method for knowledge graph completion the same as \citet{lopez2020fully}. For the input context, We trimmed the sentence to a maximum of 25 words. During the trimming, one word at a time is removed from one side of the mention, trying to keep the mention in the center of the sentence, and preserve the context information of the mention. For the input mention, we trimmed the mention to a maximum of 5 words. 

\section{Experiment Details}

All of our experiments use 32-bit floating point numbers, not 64-bit floating point numbers as in most previous work. We use 1 RTX 2080Ti for knowledge graph completion, entity typing and probing, 4 RTX 2080Ti for machine translation. We use PyTorch as the neural networks' framework. The negative curvature $K$ of the Lorentz model in our experiments is $-1$.

We take the function $\phi$ in Lorentz linear layer to have the form 
\small
\begin{equation}
    \phi(\mathbf{Wx})=\frac{\sqrt{(\lambda\sigma(\mathbf{v}^T\mathbf{x}+b)+\epsilon)^2+1/K}}{\lVert\mathbf{W\cdot\texttt{dropout}(x)}\rVert}\mathbf{W\cdot\texttt{dropout}(x)}.
    \label{eq:linear}
\end{equation}
\normalsize

To see what it means, we first compute $\mathbf{y}_0=\lambda\sigma(\mathbf{v}^T\mathbf{x}+b)+\epsilon$ as the $0$-th dimension of the output $\mathbf{y}$, where $\sigma$ is the sigmoid function, $\lambda$ controls the $0$-th dimension's range, it can be either learnable or fixed, $b$ is a learnable bias term, and $\epsilon>\sqrt{1/K}$ is a constant preventing the $0$-th dimension be smaller than $\sqrt{1/K}$. According to the definition of Lorentz model,  $\mathbf{y}$ should satisfies $\lVert\mathbf{y}_{1:n}\rVert^2-\mathbf{y_0}^2=1/K$, that is, $\lVert\mathbf{y}_{1:n}\rVert=\sqrt{\mathbf{y_0}^2+1/K}=\sqrt{(\lambda\sigma(\mathbf{v}^T\mathbf{x}+b)+\epsilon)^2+1/K}$. Then \cref{eq:linear} can be seen as first calculate $\tilde{\mathbf{y}}_{1:n}=\mathbf{W}\cdot\texttt{dropout}(\mathbf{x})$, then scale $\tilde{\mathbf{y}}_{1:n}$ to have vector norm $\lVert\mathbf{y}_{1:n}\rVert$ to obtain $\mathbf{y}_{1:n}$. Finally, we concatenate $\mathbf{y_0}$ with $\mathbf{y}_{1:n}$ as output.

In residual layer, $\phi$ is defined as
\small
\begin{equation}
    \phi(\mathbf{Wy})=\frac{\sqrt{(\mathbf{v}^T\mathbf{y}+\mathbf{x}_0+b)^2+1/K}}{\mathbf{W}\cdot\texttt{dropout}(\mathbf{y})}\mathbf{W}\cdot\texttt{dropout}(\mathbf{y}),
\end{equation}
\normalsize
where $\mathbf{x}, \mathbf{y}$ are the notations in Eq. 16.



\subsection{Initialization}

\bgroup
\setlength{\tabcolsep}{4pt}
\begin{table}[t!]
  \caption{Statistics of FB15k-237 and WN18RR.}
  \vskip 0.15in
\small
  \begin{tabular}{l r r r r r}
    \toprule
    \textbf{Dataset} & \textbf{\#Entity} & \textbf{\#Relation} & \textbf{\#Train} & \textbf{\#Valid} & \textbf{\#Test}\\
    \midrule
    FB15k-237 & 14,541 & 237 & 272,115 & 17,535 & 20,466\\
    WN18RR & 40,943 & 11 & 86,835 & 3,034 & 3,134\\
    \bottomrule
  \end{tabular}
  \label{tab:statistics-graph}
  \vskip -0.1in
\end{table}
\egroup

\begin{table}
    \setlength{\tabcolsep}{3pt}
    \caption{Initialization methods of different parameters.}
    \vskip 0.15in
    \begin{tabular}{l r}
        \toprule
        \texttt{Embedding}\\
        $\mathbf{x}_i$ & Geoopt default\\
        Parameters in $f$ & PyTorch default\\
        \midrule
        \texttt{Lorentz Linear Layer} & \\
        $\mathbf{W}$ & \texttt{Uniform}(-1/dim, 1/dim)\\
        $\mathbf{v}$ & \texttt{Uniform}(-1/dim, 1/dim)\\
        \midrule
        \texttt{Lorentz Residual}\\
        $\mathbf{W}$ & PyTorch default\\
        $\mathbf{v}$ & PyTorch default\\
        \bottomrule
    \end{tabular}
    \vskip -0.1in
    \label{tab:init}
\end{table}
We use different initialization method for different parameters, see Table~\ref{tab:init}. Geoopt\cite{geoopt2020kochurov} initialize the parameter with Gaussian distribution in the tangent space, and map the embedding to hyperbolic space with exponential map. For hyperbolic embedding of knowledge graph completion, we use a Gaussian distribution with standard deviation equals to $1/\sqrt{\texttt{dim}}$ in tangent space, for other tasks, we use a standard normal distribution.

\subsection{Knowledge Graph Completion}

\begin{table}[t!]
    \caption{Hyper-parameters for knowldge graph completion.}
    \vskip 0.15in
    \centering
    \begin{tabular}{l r r r r}
        \toprule
        & \multicolumn{2}{c}{\textbf{WN18RR}} & \multicolumn{2}{c}{\textbf{FB15k-237}}\\
        Dimension & 40 & 200 & 40 & 200\\
        \midrule
        Batch Size & 1000 & 1000 & 500 & 500\\
        Neg Samples & 4 & 1 & 64 & 64\\
        Margin & 8.0 & 8.0 & 6.0 & 6.0\\
        Epochs & 3000 & 3000 & 500 & 500\\
        Max Norm & 1.5 & 1.5 & 1.5 & 1.5\\
        $\lambda$ & 2.5 & 2.5 & 2.5 & 2.5\\
        Learning Rate & 250 & 250 & 50 & 50\\
        Optimizer & rSGD & rSGD & rSGD & rSGD\\
        \bottomrule
    \end{tabular}
    \vskip -0.1in
    \label{tab:hyp-graph}
\end{table}

We list the hyper-parameters used in the experiment in Table~\ref{tab:hyp-graph}. Note that in this experiment, we restrict the norm of the last $n$ dimension of the embeddings to be no bigger than a certain value, referred to as Max Norm in Table~\ref{tab:hyp-graph}. For each dataset, we explore $\text{BatchSize}\in\{500, 1000, 2000\}$, $\text{NegSamples}\in\{1, 2, 4, 8, 50, 64\}$, $\text{Margin}\in\{4, 6, 8\}$, $\text{MaxNorm}\in\{1.5, 2.5, 3.5\}$, $\lambda\in\{2.5, 3.5, 5.5\}$, $\text{LearningRate}\in\{250, 100, 50\}$.

\begin{table}
    \caption{Hyper-parameters for machine translation.}
    \vskip 0.15in
    \centering
    \begin{tabular}{l r r}
        \toprule
        \textbf{Hyper-parameter} & \textbf{IWSLT'14} & \textbf{WMT'16}\\
        \midrule
        GPU Numbers & 4 & 4\\
        Embedding Dimension & 64 & 64\\
        Feed-forward Dimension & 256 & 256\\
        Batch Type & Token & Token\\
        Batch Size & 3300 & 4096\\
        Gradient Accumulation Steps & 3 & 3\\
        Training Steps & 30000 & 200000\\
        Dropout & 0.0 & 0.0\\
        Attention Dropout & 0.1 & 0.1\\
        Max Gradient Norm & 0.5 & 0.5\\
        Warmup Steps & 2000 & 6000\\
        Decay Method & noam & noam\\
        Label Smoothing & 0.1 & 0.1\\
        Layer Number & 6 & 6\\
        Head Number & 6 & 6\\
        Learning Rate & 2 & 2\\
        Adam Beta2 & 0.998 & 0.998\\
        Optimizer & rAdam & rAdam\\
        \bottomrule
    \end{tabular}
    \vskip -0.1in
    \label{tab:hyp-mt}

\end{table}

\subsection{Machine Translation}
Our code is based on OpenNMT's Transformer\cite{klein-etal-2017-opennmt}. The hyper-parameters are listed in Table~\ref{tab:hyp-mt}
\subsection{Dependency Tree Probing}

The probing for the Euclidean Transformer is done by first applying an Euclidean linear mapping $f_P:\Real^n\rightarrow\Real^{m+1}$ followed by a projection to map Transformer's intermediate context-aware representation $\mathbf{c}_i$ into points $\mathbf{\tilde{h}}_i$ in tangent space of Lorentz model's origin, then using exponential map to map $\mathbf{\tilde{h}}_i$ to hyperbolic space $\mathbf{p}_i$. In the hyperbolic space, we construct the Lorentz syntactic subspace via a Lorentz linear layer $f_Q:\mathbb{L}_K^m\rightarrow\mathbb{L}_K^m$:
\begin{equation*}
\begin{aligned}
  \mathbf{p}_i &= \exp_\mathbf{0}^K(f_P(\mathbf{c}_i)),\\
  \mathbf{q}_i &= f_Q(\mathbf{p}_i).
\end{aligned}
\end{equation*}

We use the squared Lorentzian distance between $\mathbf{q}_i$ and $\mathbf{q}_j$ to recreate tree distances between word pairs $w_i$ and $w_j$, the squared Lorentzian distance between $\mathbf{q}_i$ and the origin $\mathbf{o}$ to recreate the depth of word $w_i$. We minimize the following loss:
\begin{equation*}
\begin{aligned}
  \mathcal{L}_{\text{distance}}&=\frac{1}{l^2}\sum_{i,j\in\{1,\cdots,t\}}|d_T(w_i, w_j)-d_{\mathcal{L}}^2(\mathbf{q}_i, \mathbf{q}_j)|\\
  \mathcal{L}_{\text{depth}}&=\frac{1}{l}\sum_{i\in\{1,\cdots,t\}}|d_D(w_i)-d_{\mathcal{L}}^2(\mathbf{q}_i, \mathbf{o})|,
\end{aligned}
\end{equation*}
where $d_T(w_i,w_j)$ is the edge number of the shortest path from $w_i$ to $w_j$ in the dependency tree, and $l$ is the sentence length. For the probing of Lorentz Transformer, we only substitute $f_P$ with a Lorentz one, and discard the exponential map. We probe every layer for both models, and report the results of the best layer.

We do the probing in the 64 dimensional hyperbolic space. The hyper-parameters and the best layer we choose according to development set are listed in Table~\ref{tab:hyp-probe}. Because no Lorentz embedding is involved, we simply use Adam\cite{DBLP:journals/corr/KingmaB14} as the optimizer. For parameter selection, we explore $\text{Dropout}\in\{0.0, 0.1, 0.2\}$, $\text{AttDropout}\in\{0.0, 0.1, 0.2\}$, $\text{MaxGradNorm}\in\{0.1, 0.5, 1\}$, $\text{LearningRate}\in\{2, 2.5, 3\}$

\subsection{Entity Typing}
We initialize the word embeddings by isometrically projecting the pretrained \poincare Glove word embeddings~\cite{tifrea2018poincare} to Lorentz model, and fix them during training. A Lorentz linear layer is applied to transform the word embeddings to a higher dimension. To get mention representation, the mention encoder first obtain word representation $\mathbf{s}_i$ through position encoding module described in section 3.4, then calculate the centroid of $\mathbf{s}_i$ as mention representation $\mathbf{m}_i$ according to Eq. 14 with uniform weight
\begin{equation*}
\begin{aligned}
  \mathbf{s}_i &= \texttt{PE}(\texttt{HL}(\mathbf{w}_i)),\\
  \mathbf{m}_i &= \texttt{Centroid}(1/l, \mathbf{s}_i),
\end{aligned}
\end{equation*}
where $l$ is the length of sentence, $\mathbf{w}_i$ is the pretrained embedding of $i$-th word. The context encoder is a Lorentz Transformer encoder that shares the same embedding module with mention encoder. The context representation $\mathbf{c}$ is the distance-based attention~\cite{lopez2020fully} result over the Transformer encoder's output $\mathbf{h}_i$:
\begin{equation*}
\begin{aligned}
  \mathbf{x}_i = \texttt{PE}(\mathbf{h}_i)&,\quad
  \mathbf{q}_i = \texttt{HL}(\mathbf{x}_i),\quad
  \mathbf{k}_i = \texttt{HL}(\mathbf{x}_i),\\
  \nu_{ij} &= \texttt{Softmax}(-d_{\Lc}^2(\mathbf{q}_i, \mathbf{k}_j) / \sqrt{n}),\\
  \mathbf{c} &= \texttt{Centroid}(\nu_{ij}, \mathbf{h}_i).
\end{aligned}
\end{equation*}

\begin{table}
    \caption{Hyper-parameters for dependency tree probing.}
    \vskip 0.15in
    \centering
    \begin{tabular}{l r r r}
        \toprule
        Hyper-parameter & Euclidean & HyboNet & eHyboNet\\
        \midrule
        Learning Rate & 5e-4 & 5e-4 & 1e-4\\
        Best Layer & 0 & 4 & 5\\
        Warmup steps & 1000 & 1000 & 1000\\
        Dropout & 0.0 & 0.0 & 0.0\\
        Batch Size & 32 & 32 & 32\\
        Epochs & 5 & 5 & 5\\
        Optimizer & Adam & Adam & Adam\\
        \bottomrule
    \end{tabular}
    \vskip -0.1in
    \label{tab:hyp-probe}

\end{table}

\begin{table}
    \caption{Hyper-parameters for dependency tree probing.}
    \vskip 0.15in
    \centering
    \begin{tabular}{l r}
        \toprule
        Context Encoder Layer & 2\\
        Embedding Dimension & 128\\
        Feed-forward Dimension & 512\\
        Dropout & 0\\
        Attention Dropout & 0.2\\
        Final Merge Dropout & 0.1\\
        Crowd Cycles & 10\\
        Learning Rate & 1e-3\\
        Batch Size & 900\\
        Epochs & 60\\
        Max Gradient Norm & 0.2\\
        Optimizer & rAdam\\
        \bottomrule
    \end{tabular}
    \vskip -0.1in
    \label{tab:hyp-typing}
\end{table}

To get the final representation $\mathbf{r}$, we combine $\mathbf{m}$ and $\mathbf{c}$ in the way of combining multi-headed outputs described in Eq. 16 by regarding $\mathbf{m}$ and $\mathbf{c}$ as a two-headed output. We then calculate a probability $p_i$ for every type label:
\begin{equation*}
\begin{aligned}
  p_i = \sigma(-d_\mathcal{L}^2(\mathbf{r}, \mathbf{t}_i) / \alpha_i + \beta_i),
\end{aligned}
\end{equation*}
where $\sigma$ is sigmoid function, $\mathbf{t}_i$ is the Lorentz embedding of the $i$-th type label, and $\alpha_i, \beta_i$ are learnable scale and bias factor respectively. During training, we optimize the multi-task objective~\cite{vaswani2017attention}. For evaluation, a type is predicted if its probability is larger than $0.5$.%
The hyper-parameters are listed in Table~\ref{tab:hyp-typing}. For the learnable parameters $\alpha_i$ and $\beta_i$, we initialize $\alpha_i=1$ and $\beta_i=20$. An important observation is that $\beta_i$ should be initialized around the mean value of $d_\mathcal{L}^2(\mathbf{r}, \mathbf{t}_i)$. For parameter selection, we explore $\text{Dropout}\in\{0.0, 0.1, 0.2\}$, $\text{AttDropout}\in\{0.0, 0.1, 0.2\}$, $\text{MergeDropout}\in\{0.0, 0.1\}$, $\text{CrowdCycles}\in\{5, 10, 15\}, \text{LearningRate}\in\{1e-3, 2e-3, 8e-4\}$, $\text{MaxGradNorm}$

\bibliography{icml2021}
\bibliographystyle{icml2021}